\pdfoutput=1
\documentclass[11pt]{article}
\usepackage[table]{xcolor}
\usepackage{acl}
\usepackage{times}
\usepackage{latexsym}
\usepackage{booktabs}
\usepackage[T1]{fontenc}
\usepackage{subcaption}
\usepackage{authblk}
\usepackage[utf8]{inputenc}
\usepackage{algorithm}
\usepackage{caption}
\usepackage{soul}
\usepackage{algpseudocode} 
\usepackage{amsmath} 
\usepackage{amssymb} 
\usepackage{microtype}
\usepackage{inconsolata}
\usepackage{placeins}
\usepackage{graphicx}
\usepackage[toc,page,header]{appendix}
\usepackage{minitoc}
 \usepackage{hyperref}

\algrenewcommand\algorithmicrequire{\textbf{Input:}}
\algrenewcommand\algorithmicensure{\textbf{Output:}}

\title{NewsInterview: a Dataset and a Playground to Evaluate LLMs' Grounding Gap via Informational Interviews}

\author[1*]{\bf Alexander Spangher}
\author[2*]{\bf Michael Lu}
\author[2]{\bf Sriya Kalyan}
\author[1]{\bf Hyundong Cho}
\author[1]{\bf Tenghao Huang}
\author[3]{\authorcr \textbf{Weiyan Shi}}
\author[1]{\bf Jonathan May}
\affil[1]{University of Southern California Information Sciences Institute}
\affil[2]{University of California, Berkeley}
\affil[3]{Northeastern University}
\affil[*]{\texttt{spangher@usc.edu}}

\begin{document}
\maketitle
\doparttoc 
\faketableofcontents 



\begin{figure}[t]
    \centering
    \begin{subfigure}[t]{\linewidth}
        \centering
        \includegraphics[width=\linewidth]{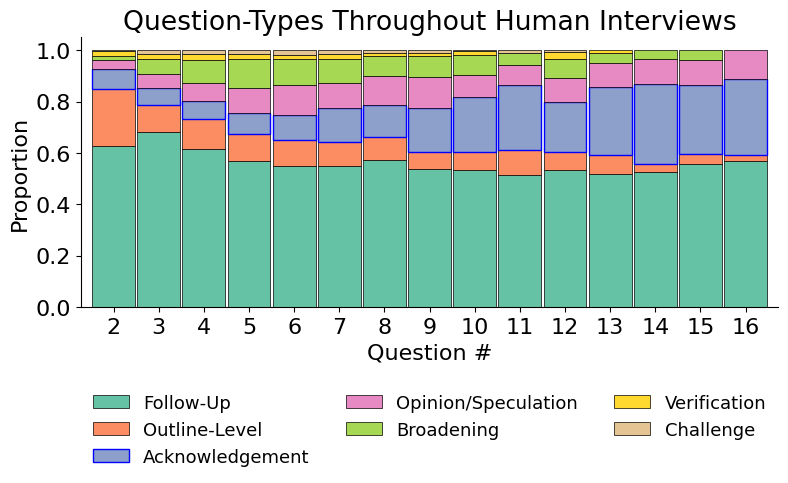}
        \caption{\textbf{Proportion of Discourse types throughout human interviews}. Human journalists use different discourse roles across the interview, including gradually more Acknowledging statements, increasing from 5\% at the start to over 20\% by the end.}
        \label{fig:human-discourse-across-time}
    \end{subfigure}
    
    \vspace{1em}  

    \begin{subfigure}[t]{\linewidth}
        \centering
        \includegraphics[width=\linewidth]{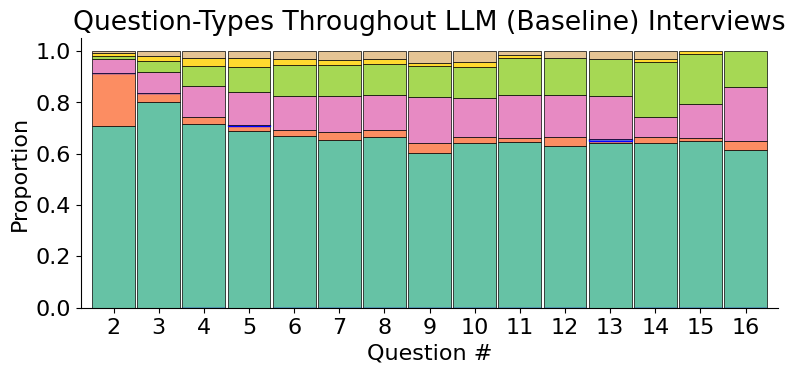}
        \caption{\textbf{Proportion of Discourse types of LLM responses in interviews}. LLMs display an increasing likelihood of asking opinion or broadening questions over the course of an interview and a lower likelihood of returning to outline-level questions.}
        \label{fig:llm-discourse-across-time}
    \end{subfigure}
    \caption{Comparison of discourse types across interviews (the first turn, usually a greeting, is excluded). The LLM is shown the first $t-1$ turns of a human interview and asked to generate the next question.}
    \label{fig:comparison-discourse-over-time}
\end{figure}

\begin{abstract}

Large Language Models (LLMs) have demonstrated impressive capabilities in generating coherent text but often struggle with strategic dialogue. To address this gap, 
we focus on journalistic interviews
. We curate a dataset of 40,000 two-person informational interviews from major news organizations in scenarios where human interviewers employ strategies to coax information from sources. We then try to mimic these activities with LLMs and find striking differences; models are less likely to use acknowledgments and more likely to rabbit-hole and not pivot to other topics. Realizing that a fundamental deficit exists in LLM multi-turn planning and strategic thinking, we develop a realistic simulated environment, incorporating source personas and persuasive elements, in order to facilitate the development of agents with long-horizon rewards. Our experiments show that mimicry failures are not two-sided; when posing as a source, models adequately reflect human behavior in information sharing, making our simulation a realistic benchmark. Interviewer-LLMs, however, struggle with engaging persuasively, leading to suboptimal information extraction across model size and capability. This simulated game 
 lays the groundwork for future work in enhancing LLMs' strategic dialogue capabilities.\footnote{We release dataset and code at \url{https://github.com/alex2awesome/news-interview-question-generation}.} 

\end{abstract}

\section{Introduction}



Recent research has shown that LLMs struggle to engage in emotional \cite{shaikh2024grounding} or strategic \cite{wongkamjan2024more} dialogue. For example, \newcite{shaikh2024grounding} examined LLM-generated responses to dialogues and found fewer occurrences of ``grounding language''~\cite{clark1996using, cho2020grounding}, like acknowledgements or affirmations, that humans typically use to foster comfort and trust. This can impede an LLM's ability to serve in a variety of situations: e.g., education \cite{kasneci2023chatgpt}, mental health \cite{carlbring2023new} or conflict resolution \cite{argyle2023ai}. However, prior efforts to ameliorate such gaps face limitations: existing large datasets (1k--10k transcripts) are generated via crowdsourcing and are inherently unnatural \cite{rashkin-etal-2019-towards, wang2019persuasion,liu2021towards}. More natural datasets, of educational \cite{caines2020teacher} or therapeutic environments  \cite{gratch-etal-2014-distress}, are difficult to collect due to privacy concerns \cite{casey2004challenges} and are small-scale (100--1k transcripts). 

In this work, we directly address these limitations by focusing on an area where grounding communication is required but plentiful data exist: journalist interviews. Journalistic, or \textit{informational} interviews, are typically conducted between an ``interviewer'' and a ``source,'' and the goal is to obtain information. 
Sources are often anxious or unclear \cite{harcup2015journalism}, and human interviewers are constantly evaluating: (1) Are my questions getting fully addressed? (2) Do I need to more effectively engage or persuade a source \cite{sedorkin2015interviewing}? This makes news interviews an ideal setting to observe grounding dialogues.

To study how LLMs perform in journalistic contexts, we start by collecting interview transcripts from two major US news sources: National Public Radio (NPR) and Cable News Network (CNN), filtering to over 40,000 dyadic informational interviews.\footnote{As opposed to games, questionaires and other formats these news outlets release} Next, we show that LLMs in news interview settings suffer from the same lack of grounding as in other dialogue settings \cite{shaikh2024grounding}. We find that significant discourse differences exist in the kinds of questions asked by LLMs: for example, LLMs are 50\% less likely to make acknowledgements, and 30\% less likely to pivot to higher-level questions.

Next, we turn to a more fundamental question: \textit{why consider grounding?} According to \newcite{cialdini2009influence}, grounding exists for a purpose: to influence an outcome (e.g., in therapeutic environments, the grounded patient is more open and makes more progress \cite{bohart1999clients}; in educational environments, the comfortable student is more engaged and learns more \cite{brown1994guided}; these are all usually borne out over many conversational turns). Observing the \textit{effects} of grounding language in terms of objective outcomes might be a more effective way to reason and ultimately train empathetic agents and improve long-horizon strategic dialogue.
Motivated by these observations, we develop a realistic game environment to serve as a playground: in this simulation, LLMs play the role of the interviewer and the source. The goal for the interviewer is to \textit{obtain the maximal amount of information from the source in a limited number of questions}.

In order to induce the need for grounding communication, we design different personas for sources (e.g., anxious, clueless, dominating), each with different communication patterns. We also add a responsiveness to strategic dialogue: sources will only return information if they are persuaded in a manner befitting their personas\footnote{We understand that ``being persuaded,'' ``being made comfortable,'' and ``being acknowledged'' are all separate forms of grounding, some more active than others. However, we use ``persuasion'' as a short-hand encompassing all categories.} \cite{harcup2015journalism, sedorkin2015interviewing}. 
We find that our environment is realistic: source-LLMs correlate significantly with humans in their ability to identify persuasion ($r=.43$, $p<.0001$). However, interviewer-LLMs struggle to both recognize when questions are answered and actively persuade the source, resulting in suboptimal information extraction.

In summary, our contributions are: 
\begin{itemize}
\item We release a high-quality dataset of 40,000 dyadic informational interviews from NPR and CNN. This dataset addresses the scarcity of large-scale dialogue data necessary for studying grounding communication.

\item We perform a detailed discourse analysis comparing LLM-generated dialogues with human interviewers, identifying significant differences in the use of grounding language and question types.

\item We develop a game environment to test and improve dialogue agents in informational interviews, which we call \textit{NewsInterview}. Our findings indicate this is a realistic setting but highlight the challenges LLM interviewers face in engaging in persuasive dialogue.
\end{itemize}

\section{Dataset Processing}

\subsection{Data Collection}

We aggregate, clean and condense multiple publicly available datasets of interview transcripts from NPR and CNN in order to build a high-quality interview dataset of 45k source-to-interview transcripts. These transcripts are published records of live interviews conducted between a journalist and sources invited on the program. They provide a rich resource for analyzing natural language interactions.

\subsection{Data Filtering for Interview Analysis}

We want to focus on one-on-one informational interviews between a journalist and a single source. We start with 487,310 transcripts collected by \citet{majumder2020interview} and \citet{zhu2021mediasum}. However, initial examination of the transcripts reveals many of them to be low-quality: they include multiple sources, are formatted as panel discussions, or are not informational in nature (e.g., they include game shows). To filter the transcripts and retain only those that fit our criteria, we prompt Llama-3.1-70b\footnote{\url{https://huggingface.co/meta-llama/Meta-Llama-3.1-70B-Instruct} \cite{touvron2023llama} using the vLLM framework \cite{kwon2023efficient}} to classify each transcript based on the number of participants and the nature of the content. The prompts used for filtering are provided in App. \ref{app:prompts}. After filtering,  45,848 interviews remain. Finally, the original transcripts do not distinguish which participant was the interviewer vs. the interviewee. So, we count each participant's use of question marks: the participant with more is labeled the interviewer.\footnote{Manual validation on 50 interviews showed this method correctly identified roles in $>98\%$ of cases.}

Conversations have, on average, 7.5 turns between the interviewer and source. The source speaks for longer, with an average of 551 words per conversation compared with the interviewer's 270 words (or 27 words per source utterance, 16 per interviewer). Interviewers tend to ask ``what'' and ``how'' questions the most, and conversations occur at Flesch-Kincaid Grade of 6.9  \cite{kincaid1975derivation}. Interviews cover a range of topics, from \colorbox{red!10}{literature}, \colorbox{blue!10}{politics}, \colorbox{green!10}{academics}, and \colorbox{yellow!10}{international affairs} (see Appendix \ref{app:eda} for more details).


\begin{table*}[t]
    \centering
    \begin{tabular}{lrrrrrr}
    \toprule
     & Exact Match & Info. & Motivation & Style & Discourse & Context \\
    \midrule
    Baseline-LLM & 3.9\% & 4.4\% & 4.7\% & 11.9\% & 36.2\% & 53.0\% \\
    Chain-of-Thought (CoT) & 4.5\% & 3.6\% & 5.2\% & 12.8\% & 37.0\% & 56.9\% \\
    LLM w. Outline & 3.7\% & 3.8\% & 4.1\% & 9.6\% & 36.2\% & 46.6\% \\
    Outline-CoT & 3.6\% &    3.9\% &    4.3\% &   8.3\% & 29.9\% & 43.1\% \\
    \midrule
    Human       & 8.2\% & 17.5\%  & 35.4\%   & 40.2\%  & 54.5\%   & 60.3\%  \\
    \bottomrule
    \end{tabular}
    \caption{\textbf{Alignment of LLM-Generated Questions with Original Interview questions}. We give an LLM, \texttt{Llama-3.1-70b}, the prior $t-1$ turns in an interview and prompt it to ask the next question. We measure the percentage of times this question aligns to a question asked by a human at the same point in the interview across six dimensions: Exact Match (nearly exactly the same as the original utterance), Information (relevant factual content), Motivation (same motivation as the original question), Style (alignment with tone and phrasing), Discourse (structural role within the interview), and Context (incorporation of contextual knowledge). The prompting strategies compared are Baseline-LLM, Chain-of-Thought (CoT), LLM with an Outline, and Outline-CoT; and, we conduct a human baseline trial with a former professional journalist.}
    \label{tab:consistency_evaluation}
\end{table*}

\begin{figure*}[t]
    \centering
    \includegraphics[width=.8\linewidth]{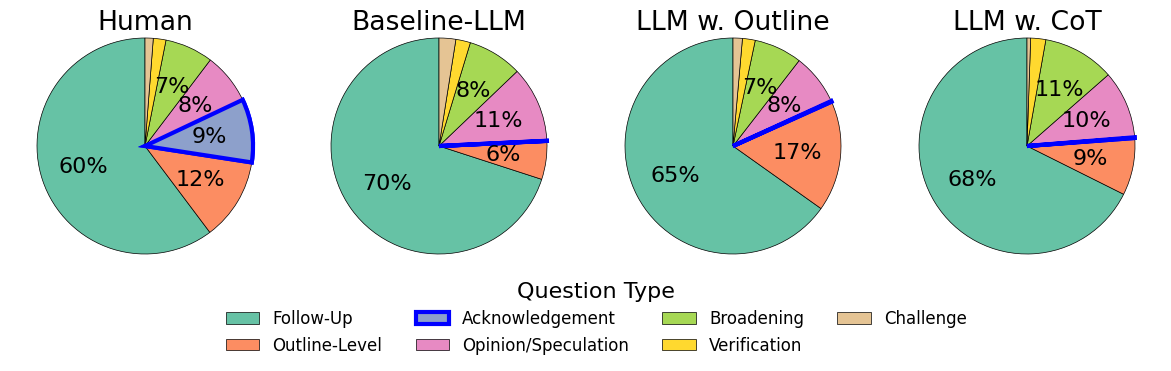}
    \caption{\textbf{Distribution of Discourse Roles in Questions, Across Different Prompting Strategies}. We compare the proportions of discourse roles of questions (e.g., ``Follow-up,'' ``Acknowledgement,'') generated by (a) human journalists, (b) Baseline-LLM (\texttt{Llama-3.1-70b}) (c) LLM prompted with an Outline and (d) with Chain-of-Thought (CoT). Acknowledgement statements, which often build empathy, are significantly underrepresented in all LLM prompting approaches, compared to human-generated questions (see appendix for Outline-CoT).}
    \label{fig:discoure_pie_charts}
\end{figure*}

\section{Analysis}

In this section, we analyze how humans conduct informational interviews and compare this behavior to that of pretrained LLMs, to explore whether LLMs face similar grounding problems as observed in other settings \cite{clark1996using, shaikh-etal-2024-grounding}.

\subsection{Generating Counterfactual Utterances}

One way to assess how an LLM would behave in an interview setting offline is to perform a counterfactual simulation \cite{shaikh-etal-2024-grounding}. Specifically, given a human interview consisting of at least $t$ interviewer-source conversational turns $(q_1, a_1)...(q_t, a_t)...$, we feed $t-1$ turns into the LLM along with a prompt instructing the LLM to generate the next question. This generates a counterfactual, $g_t$ to what the human would have said, $q_t$. We experiment with different variations: (1) \textbf{Baseline}: The LLM is simply asked to produce the next question. (2) \textbf{Chain-of-Thought (CoT)}: The LLM is instructed to reason about the information already provided in the interview, consider what might be left to ask, and then generate the next question. (3) \textbf{Outline}: the LLM is provided with an outline of the interview goals (described in Section \ref{sct:dataset_preparation}) to incorporate into CoT reasoning.\footnote{We include full prompt examples for all three variations in Appendix \ref{app:prompts}. All question-generation experiments are conducted using \texttt{Llama-3.1-70b}.}

\subsection{Evaluating LLM Counterfactuals}

To analyze how similar LLM questions, $g_t$ are to human questions, $q_t$, we perform two analyses:

\paragraph{Consistency Analysis:} We aim to assess how similar $g_t$ is to $q_t$ across different comparison categories \newcite{saha2024branch}, specifically: \textit{Informational} consistency (i.e., $g_t$ and $q_t$ seek similar informational objectives); \textit{Motivational}, (i.e., similar outcomes); \textit{Style}, (i.e., similar tone); \textit{Contextual} consistency (i.e., similar appropriateness given the context); \textit{Discourse} consistency (i.e., similar purposes in the overall conversation). Putting these together, we assess an $Exact$ match. We ask an LLM, \texttt{GPT-4o}, to perform this assessment and manually inspect its outputs and reasoning threads.

\paragraph{Discourse Analysis:} We aim to assess whether $g_t$ plays a similar \textit{function} as $q_t$ does. We develop a schema to describe the role of each question.\footnote{To generate our discourse schema, we asked two journalists to analyze fifty interview transcripts. One had eight years of experience in newsrooms, the other was an undergraduate student studying journalism. We held three conferencing sessions to develop the schema. Then, we blindly annotated ten interviews, achieving a $\kappa=.6$. Given our schema, we then asked an LLM to classify discourse roles in sentences. The prompt contains the interview context, $(q_1, a_1)...(q_{t-1}, a_{t-1})$, and current question $q_t$. To validate the LLM's labeling accuracy, we had the professional journalist label 10 additional interviews as ground-truth and scored the LLM's assignments. The LLM scored a $.8$ f1 score.} This schema includes the following elements: \textit{Follow-up Question} (e.g., ``Can you tell us more?''), \textit{Outline-Level Question} (e.g., ``Moving on, can we discuss the next event?''), \textit{Acknowledgement Statement} (e.g., ``I see, that sounds scary.''), \textit{Opinion/Speculation} (e.g., ``What do you think will happen?''), \textit{Broadening Question} (e.g., ``How does this fit into the broader trend?''), \textit{Verification Question} (e.g., ``So to confirm...'') and \textit{Challenge Question} (e.g., ``These dates don't line up.''). See Table~\ref{tab:discourse-types} in the Appendix for definitions of each role.

\subsection{Findings}
\label{sct:analysis_results}


\paragraph{Insight \#1: Acknowledgement statements are virtually absent from all LLM variations.} As shown in Figure \ref{fig:discoure_pie_charts}, grounding gaps exist in journalistic interviewing similar to those observed by \citet{shaikh2024grounding}. While human journalistic interviewers tend to make Acknowledgement statements in about 9\% of their utterances, all prompting variations that we experimented with made close to zero of these statements. This lack of acknowledgement is paired with not mirroring the source's speaking style; human journalists, as shown in Appendix \ref{app:interview_examples}, bring character and voice. 

\paragraph{Insight \#2: LLMs do not engage in strategic multi-turn questioning.} Even in settings where LLMs are exposed to interview outlines, they are still undirected in their questions. As shown in Figure \ref{fig:discoure_pie_charts}, LLMs are significantly more likely to ask follow-up questions than humans across all prompting variations. Introducing chain-of-thought and outline variations increases the rate at which the LLM asks outline-level questions. However, the rate remains significantly below human levels. Additionally, they are also more likely to ask either Opinion questions or Broadening questions. In fact, in Figure \ref{fig:llm-discourse-across-time}, we observe that LLMs tend to ask increasing amounts of Opinion Questions and Broadening Questions over time, which humans do not. As shown in Table \ref{tab:llm_examples}, these questions can be vague and open-ended. Together, these findings suggest an inability to direct an interview in a desired direction and engage in multi-turn planning.

\paragraph{Insight \#3: LLMs are capable of understanding context, but fail in other categories of similarity to humans.} Comparing the content and style of LLM interviews to human interviews in Table \ref{tab:consistency_evaluation}, we note that, overall, LLMs are broadly dissimilar to humans in style, motivation and information-seeking. One area where the LLMs succeed, relatively, is understanding the context of the interview beforehand. This is not a new observation -- much recent work, e.g., in dialogue-tracking, has found LLMs to perform well \cite{ou2024dialogbench}. The fact that LLMs can preserve context over multiple turns and do not drift away from the topic indicates that models might \textit{one day} be able to engage in multi-turn goal-oriented dialogue, given the right reward signals and learning environment.

Taken together, these findings suggest that journalistic dialogue is suitable for studying effective communication patterns, and also highlight significant gaps in current language modeling objectives. While LLMs can generate contextually relevant questions, they lack both an emotional and connective drive as well as the strategic planning exhibited by human interviewers. 

\begin{algorithm}[t]
\caption{Gameplay}
\label{alg:gameplay}
\hspace*{\algorithmicindent} \textbf{Input} Interviewer objectives $o$, Source Informational Items $I$, Source persona $\phi$, $K$ turns \\
\hspace*{\algorithmicindent} \textbf{Output} Reward \( R \)
\begin{algorithmic}[1]
    \State \textbf{Initialize:} Reward \( R \leftarrow 0 \), Conversation History $C\leftarrow \left[\right]$, Used items $U\leftarrow\{\}$
    \For{$i \in 1,...K$}
        \Statex  \(\triangleright\) \textbf{Step 1: Interviewer Question Generation }
        \State $g_i =$ \texttt{Interviewer}$(C, o)$
        \Statex  \(\triangleright\) \textbf{Step 2: Source's Response Generation}
        \State $r=$\texttt{getRelevantInfoItems}$(I, U, g_i)$
        \State $p=$\texttt{getPersuasionLevel}$(C)$
        \State $f=$\texttt{getItemsToReturn}$(r, p)$
        \State $a_i=$\texttt{Source}$(g_i,C,f,p,\phi)$
        \Statex  \(\triangleright\) \textbf{Update Variables}
        \State $U \leftarrow U \cup f$, $C \leftarrow C \oplus \left[g_i, a_i\right]$, $R\leftarrow R + |f|$
    \EndFor 
\end{algorithmic}
\end{algorithm}
\begin{figure*}
    \centering
    \includegraphics[width=.8\linewidth]{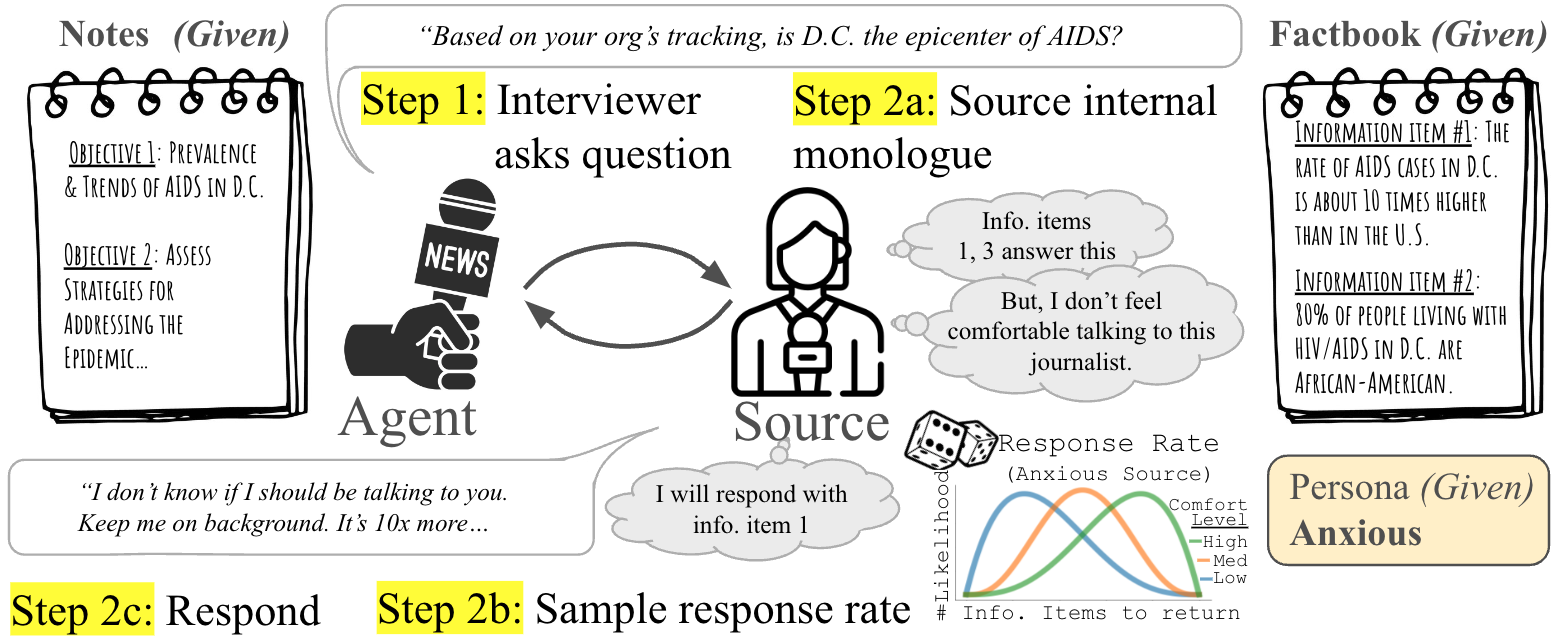}
    \caption{\textbf{Walkthrough of the LLM Interviewer-Agent Process}. In the NewsInterview game, an interviewer-LLM converses with a source-LLM: the interviewer-LLM is rewarded based on how many information items (shown at the right) are extracted from the source. In more detail: the interviewer agent is given a set of high-level objectives, similar to a journalist's pre-interview notes, while the source is given a persona and a set of relevant facts. The interview proceeds for $k$ turns. \textbf{Interviewer Query:} the interviewer is prompted to ask a question based on their goals and information obtained (Step 1). \textbf{Source Response:} The source responds with a multi-step process. First, they are prompted to determine how many information items in their factbook are relevant to the question (Step 2a). Then, they self-assess their comfort level. Depending on this, the simulation randomly selects a subset of relevant information for the response (Step 2b). \textit{We track the decision of which items to return on the back-end, in order to calculate the final reward.} The source is then prompted to craft a reply aligned with their persona (Step 2c). \textbf{After $k$ turns:} a reward given to the interviewer based on the number of information items extracted from the source.}
    \label{fig:splash_image}
\end{figure*}

\section{NewsInterview: An Interview Game}

As shown, LLM counterfactual questions exhibit several shortcomings: they are less likely to acknowledge the interviewee and focus excessively on follow-up questions. But do both of these shortcomings point to a lack of strategic multi-turn planning? In human dialogue, grounding exists for long-term strategic purposes \cite{cialdini2009influence}, yet there currently exists no way to way to obtain these kinds of long-term rewards during LLM training. Motivated by this insight, our goal for the remainder of the paper is to create and validate a realistic game-environment with a delayed reward signal. We leave to future work utilization of this framework for improving strategic dialogue.

\subsection{Game Design Overview}

We first introduce our game on a high level, illustrated in Figure~\ref{fig:splash_image}, and then describe our implementation. Our gameplay proceeds in a loop, shown in Algorithm \ref{alg:gameplay}. The ``player'' in our game plays the role of an interviewer and is able to ask questions to a source, based on the conversational history and the interview objectives (the \texttt{Interviewer()} step). The source is given a set of informational items and assesses whether any of these items are relevant to the question (the \texttt{getRelevantInfoItems()} step); the source then decides how persuaded or comfortable they are based on the conversational history (the \texttt{getPersuasionLevel()} step). Based on this, we determine the subset of relevant items the source returns (the \texttt{getItemsToReturn()}), and track these on the back-end as an accumulating reward. The reward, obtained at the end of the game, is the unique number of information items disclosed. 

\subsection{Gameplay Design}
\label{sct:gameplay_design}

To design our game, we draw heavily on two journalism textbooks: \textit{Interviewing: A Guide for Journalists and Writers}, which explains how to conduct effective interviews and speak to reluctant, defensive, or poor-explaining
sources \cite{sedorkin2015interviewing}; and \textit{Journalism: Principles and Practice}, which describes how to build trust \cite{harcup2015journalism}. We first start by describing our data processing, and then we will describe Algorithm \ref{alg:gameplay} in more detail. For all gameplay prompts, see Appendix \ref{app:gameplay_prompts}.

\paragraph{Dataset Preparation for Simulation}
\label{sct:dataset_preparation}

To prepare our dataset for use in the simulated game environment, we group together: (1) source responses and ask an LLM.\footnote{Llama-3.1-70b} to summarize a set of \textit{specific informational items} and (2) interviewer questions and ask an LLM to summarize them into a set of \textit{high-level objectives}. The sources' informational items mimic the knowledge a source likely had going into the interview\footnote{Manual evaluation confirms these information items are  present in initial interviews and are non-overlapping.} and the interviewer's objectives represent the agendas they had prior to the conversation.\footnote{Manual validation with professional journalists confirms that these outlines reasonably capture what a journalist might prepare before an interview and do not leak information.} Both of these summaries are represented in Figure \ref{fig:splash_image} as \textit{Given}, and are designed to give the interviewer-LLM and the source-LLM a basis for communication. For further examples of both,  see Tables   \ref{tab:outline_objectives} and \ref{table:info_items}in the Appendix.

\paragraph{Source Design Element \#1: Personas}

Now, we introduce the design of the source. We focus attention on this construction to build a robust game environment that accurately mimics human interactions. To make gameplay varied and challenging, we draw from \citet{sedorkin2015interviewing} to design eight different personas: \textit{Anxious}, \textit{Avoidant}, \textit{Adversarial}, \textit{Defensive}, \textit{Straightforward}, \textit{Poor Explainer}, \textit{Dominating} and \textit{Clueless}. For descriptions of each persona, as well as example responses, see Table \ref{table:personas}. These personas allow us to study how interviewers perform in a wider array of challenging scenarios.

\paragraph{Source Design Element \#2: Persuasion} The following three functions, in sequence, power our gameplay: $\texttt{getRelevantInfoItems} \rightarrow \texttt{getPersuasionLevel} \rightarrow \texttt{getItemsToReturn}$. The first, \texttt{getRelevantInfoItems}, takes the interviewer's question and determines which of the sources' information items are most relevant; it is simply a retrieval function that we implement using an LLM. \texttt{getPersuasionLevel} is a function that determines the selected source's level of comfort or persuasion (on a five point scale) in the current conversation. \texttt{getItemsToReturn} is a stochastic engine: it randomly selects, based on the persuasion level, the number of relevant information items to return: \textit{the more persuaded a source is, the more likely they are to return more information.} The persuadability component to our gameplay increases the multi-turn strategy: because persuasion is assessed with reference to the entire interview, the interviewer gets more reward for spending words \textit{early} in the interview persuading the source to feel comfortable. 

Is it sound for the source-LLM to assess its own level of persuasion? As recent research has found, LLMs are poor detectors of when they are being persuaded \cite{sakurai-miyao-2024-evaluating} and can even unknowingly persuade themselves \cite{zeng2024johnny}. Furthermore, persuadability varies from person to person \cite{wang2019persuasion,hirsh2012personalized}. Luckily, source-persuasion is a well-studied field in journalism. As a starting point, we draw from \citet{sedorkin2015interviewing}, and carefully design prompts asking an LLM to rate the persuasiveness of a prior conversation. Different source personas, according to \citet{sedorkin2015interviewing}, are persuaded by different communication patterns: e.g., \textit{Anxious} sources are distrustful of journalists; they are usually persuaded by phrases like ``I will be as fair as possible.'' We validate this in Section \ref{sct:gameplay_validation}. 

\paragraph{Source and Interviewer Responses} Based on the assessed persuasion level (1--5) of the conversation, we implement \texttt{getItemsToReturn}. This function takes in all relevant information items and randomly draws from a Beta distribution to determine what percentage of relevant information items to return. We choose five different parameterizations per persona, each corresponding to a different persuasion level. As can be seen in Figure \ref{fig:splash_image}, we choose these parameterizations such that the more persuaded a source is, the more left-skewed the distribution is. Each persona has a slightly different parameterization, reflecting that some personas need less persuasion (e.g., ``Dominant'') while others do not drastically change how much information they return even with more persuasion (e.g., poor explainer). See Figures \ref{fig:beta_distributions_a} and \ref{fig:beta_distributions_b}  in App. for the Beta distributions for each source. 

\subsection{Gameplay Validation}
\label{sct:gameplay_validation}

We conducted human trials to validate how well our gameplay environment approximates real interviews, focusing on \emph{persuasion} as a pivotal dimension. Five participants, including two professional journalists and one journalism student, each served as the ``source,'' rating their own persuasion levels turn-by-turn on a five-point scale across 72 trials (576 turns total). The game's LLM-based source also generated persuasion estimates. We found a moderate but significant correlation of \(r=0.43\) (p < .0001). Excluding adversarial personas, correlation rose to \(r=0.68\). Bootstrapped estimates confirmed the consistency of these results, and a power analysis following guidelines from \newcite{cardjurafsky2020} showed our sample size was adequate to detect this effect.

These trials center on persuasion because the other components of our source design (i.e., retrieval of correct informational items), while crucial, leverage prior, well-studied phenomena in retrieval-augmented LLMs and prompt engineering~\cite{lewis2020retrievalqa, cookbooksample}. Our environment reuses standard cross-encoder reranking and chain-of-thought prompts \cite{chainofthought, reimers2019sentence}, meaning that the correct factual content is generally well-handled without substantial new techniques. Minimal forms of self-reflection~\cite{shinn2023reflexion} were used to mitigate hallucinations, and no significant factual drift was observed. Hallucinations are well-studied in the literature \cite{huang2025survey}.

Taken together, this validation suggests that modeling source \emph{persuadability} in a turn-level simulation is reasonably accurate and stable. By capturing how LLMs adapt their strategies across different personas and persuasion thresholds, our system can potentially serve as a stepping stone for training more sophisticated interview agents or supporting journalism students. Future work might expand the environment's human trials, repeat experiments at larger scale, and incorporate further realism checks to ensure robust dialogue performance and fidelity.

\begin{table*}[t]
    \centering
    \begin{tabular}{p{3.5cm}p{3.5cm}p{3.5cm}p{3.5cm}}
    \toprule
    {}    &  \textbf{Hardest} & \textbf{Medium}    & \textbf{Easiest} \\
    Model & Full Game & \textit{sans.} Persuasion & \textit{sans.} Info. witholding \\
    \midrule
    gpt-4o-mini & 49.3\% & 47.5\% & 84.7\% \\
    gpt-4o & 50.4\% & 49.8\% & 84.2\% \\
    Llama-3.1-70b & 42.6\% & 45.5\% & 80.1\% \\
    Llama-3.1-8b & 42.4\% & 48.3\% & 74.9\% \\
    \bottomrule
    \end{tabular}
    \caption{\textbf{Performance of LLMs as Interviewers, with Ablations} Percentage of information items extracted (Reward percentage) in each interview by different language models (gpt-4o-mini, gpt-4o, Llama-3.1-70b, and Llama-3.1-8b) across three conditions: (1) \textbf{Hardest}: The full game, with information dependent on persuasion and persona.  (2) \textbf{Medium}: an ablation removing the sources' responsiveness to persuasion. (3) \textbf{Easy}: An ablation removing the random withholding of information (i.e., a source returns all relevant information items at each turn). We observe, perhaps unsurprisingly, that removing the source's ability to withhold information (Medium $\rightarrow$ Easy) drastically increases the reward percentage at the end of the game. The removal of persuasion strategies has a smaller effect, with some models showing marginal gains (e.g., Llama-3.1-8b) and others slight losses (e.g., gpt-4o). This indicates that vanilla LLMs are poorly suited to this persuasion task.}
    \label{tab:results_of_gameplay}
\end{table*}

\begin{figure}[t]
    \centering
    \begin{subfigure}[t]{\linewidth}
        \centering
        \includegraphics[width=.8\linewidth]{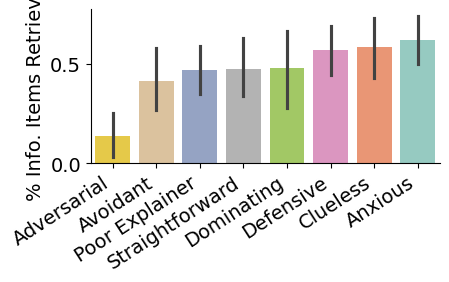}
        \caption{Rewards of \texttt{gpt-4o} from playing against sources of different persona types.}
        \label{fig:extractions-by-persona}
    \end{subfigure}
    
    \vspace{1em} 

    \begin{subfigure}[t]{\linewidth}
        \centering
        \includegraphics[width=.8\linewidth]{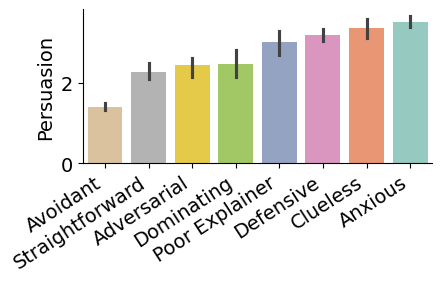}
        \caption{Average level of persuasion, from \texttt{gpt-4o}, towards the different persona types in our evaluation.}
        \label{fig:average-persuasion-by-persona}
    \end{subfigure}

    \caption{Comparison of \texttt{gpt-4o}'s performance across different persona types. The Adversarial type is by far the hardest to extract information from, however, it is easier to persuade. LLMs might be most the thrown off by adversarial sources.}
    \label{fig:combined-figures-persona}
\end{figure}

\begin{figure}[t]
    \centering
    \begin{subfigure}[t]{0.8\linewidth}
        \centering
        \includegraphics[width=\linewidth]{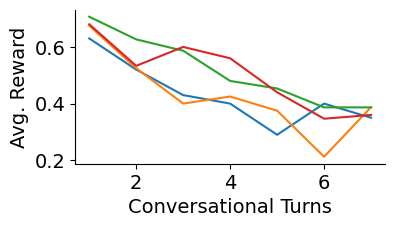}
        \caption{Average reward across conversational turns.}
        \label{fig:extractions-over-time}
    \end{subfigure}
    \vspace{1em} 
    \begin{subfigure}[t]{0.8\linewidth}
        \centering
        \includegraphics[width=\linewidth]{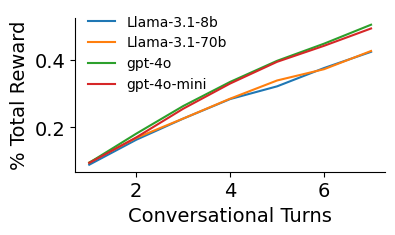}
        \caption{Percentage (\%) of Reward, by total reward.}
        \label{fig:total-extractions-over-time}
    \end{subfigure}
    \caption{
    Comparison of Rewards over time for language models. For all language models, the reward declines over time, shown above. However, this is not due to interviewer ``maxing out'' reward, as Total Reward increases nearly linearly across conversational turns.}
    \label{fig:combined-figures}
\end{figure}

\subsection{Game Simulation Results}
\label{sct:results}

We run our simulation for 450 interviews with four LLMs as the interviewer\footnote{\texttt{gpt-4o}, \texttt{gpt-4o-mini}, \texttt{Llama-3.1-70b} and \texttt{Llama-3.1-8b}} and \texttt{gpt-4o} for the source-LLM across all personas. Table \ref{tab:results_of_gameplay} compares the performance of LLMs across three conditions: the full game, a version without persuasion, and a version where sources do not withhold information. In the full game, where sources' responsiveness depends on persuasion and persona, the gpt-4o model performs the best, at 50.4\%. However, when persuasion is removed, performance only marginally improves across all models (e.g., Llama-3.1-70b reaches 45.5\%, while gpt-4o remains stable at 49.8\%), indicating that other aspects of the game (i.e., inferring which information the source has witheld) also pose a challenge. In the easiest condition, where no information withholding occurs, all models perform significantly better, with reward percentages reaching over 80\%, showing that withholding is a major obstacle.

Figure \ref{fig:extractions-by-persona} highlights the performance of gpt-4o across different source personas. The model achieves the highest information extraction from straightforward personas, while adversarial and defensive personas are the most challenging. Despite being harder to extract information from, adversarial sources are easier to persuade (Figure \ref{fig:average-persuasion-by-persona}).

Figure \ref{fig:extractions-over-time} explores how the reward (information extraction) changes over the course of an interview. The results show a declining trend in reward per conversational turn. However, the total reward accumulated over time (Figure \ref{fig:total-extractions-over-time}) increasesalmost linearly, showing that the LLMs continue to extract information, albeit at a slower rate. Together, these findings highlight the limitations of current LLMs in engaging with persuasive and strategic multi-turn interviews. While larger models like gpt-4o outperform smaller ones, they still exhibit significant gaps in persuasion and adaptive questioning, particularly when dealing with difficult personas.

\section{Discussion}

Our findings indicate that news interview transcripts provide a powerful, real-world resource for studying \emph{persuasive, grounding, and multi-turn strategies} in dialogue systems. In particular, we build on prior work that highlights grounding gaps in large language models (LLMs)~\cite{shaikh2024grounding}, extending insights from gameplay-inspired multi-turn dialogue research~\cite{wongkamjan2024more, liu2024interintent} into a domain abundant with authentic data. By examining human interviewers' behaviors, we illustrate how grounding and persuasion manifest naturally in real-world news interviews, yet remain difficult for current LLMs.

We show in Section~\ref{sct:analysis_results} that humans consistently employ grounding dialogue throughout their interviews, a tactic LLMs fail to emulate effectively. In Section \ref{sct:results}, we demonstrate how LLMs struggle to extract information from diverse source personas, particularly when those personas exhibit adversarial or avoidant traits. These findings underscore the significance of persona mismatches: while existing game-based dialogue studies often assume a single persona per environment~\cite{chawla2021casino, liu2024interintent}, our results suggest that personae with different levels of hostility or indifference pose unique challenges for current models.

One way to address these limitations is to incorporate \emph{long-range reward signals} during model training~\cite{li2016deep}. Grounding dialogue and persuasion are inherently long-horizon phenomena~\cite{clark1996using, cialdini2009influence}. In contexts like therapy, for instance, effective grounding fosters patient openness and lasting progress~\cite{bohart1999clients}; in education, it encourages students' sustained engagement and deeper learning~\cite{brown1994guided}. Our \textit{NewsInterview} framework addresses this by providing an environment in which LLMs must continually strategize about which questions to ask, what information gaps need filling, and how to persuade sources to disclose details. This game-playing setting is still less complex than fully adversarial multi-agent domains~\cite{chawla2021casino, liu2024interintent, wongkamjan2024more} because the source's goal is not to mislead but to selectively withhold information. Yet, even in this simplified scenario, LLMs struggle to maintain effective information extraction over multiple turns, pointing to deeper issues in question-asking. Future directions include refining our \texttt{getPersuasionLevel} function, introducing importance-weighted or quote-centric reward signals, and further validation.

\section{Related Work}

Research on large language models (LLMs) has increasingly underscored the importance of \emph{grounding language} and \emph{strategic dialogue}~\cite{shaikh2024grounding, wongkamjan2024more}, especially in educational, mental health, and conflict resolution settings~\cite{clark1996using, cho2020grounding, kasneci2023chatgpt, carlbring2023new, argyle2023ai}. However, current datasets for studying these phenomena are either crowdsourced and thus somewhat contrived~\cite{rashkin-etal-2019-towards, wang2019persuasion, liu2021towards}, or too small to capture the complexity of real-world interactions due to privacy constraints~\cite{gratch-etal-2014-distress, casey2004challenges, caines2020teacher}. This lack of large-scale, naturalistic data limits progress on LLMs that effectively employ affirmations and acknowledgments to establish rapport and longer-term strategic thinking~\cite{kasneci2023chatgpt}.

Meanwhile, recent research has explored the use of simulated or game-based environments for training dialogue agents on multi-turn strategic planning~\cite{lewis2017deal, he2018decoupling, gray2020human, perolat2022mastering, chawla2021casino}. Though these environments can foster negotiation and strategy, they often rely on narrowly-defined tasks or singular personas. Consequently, there is still a need for \emph{broader} conversational contexts and \emph{larger, more naturalistic} datasets~\cite{liu2021towards}. We address these gaps by introducing 40,000 two-person informational interviews from NPR and CNN, contributing a new large-scale dataset for studying grounding communication and strategic dialogue in realistic settings. Additionally, our \textit{NewsInterview} simulation environment incorporates source personas and persuasive tactics, further advancing the development and evaluation of LLMs in long-horizon, multi-turn dialogues.

\section{Conclusion}

In this paper, we have introduced a high-quality dataset of 40,000 two-person informational interviews from NPR and CNN, addressing the scarcity of large-scale dialogue data necessary for studying grounding communication. Our detailed discourse analysis reveals significant differences between LLM-generated dialogues and human interviewers, particularly in the use of grounding language and question types. Motivated by observation that long-term objectives guide turn-level grounding, we develop a realistic game environment, \textit{NewsInterview}, to test and improve dialogue agents in informational interviews. Our experiments demonstrate that while source-LLMs can mimic human behavior in information sharing, interviewer-LLMs struggle with recognizing when questions are answered and engaging persuasively, leading to suboptimal information extraction. These findings underscore the need for enhancing LLMs' strategic dialogue capabilities.

\section{Limitations}

\subsection{Privacy and Ethical Considerations}

All data used in this study are publicly available and do not contain personally identifiable information beyond what has been already made public by the news organizations. We adhere to ethical guidelines for data use and ensure that our processing respects the rights and privacy of individuals involved as well as the news organizations that collected this data. Since the dataset we create is derived from interviews that have already been published in academic settings, we believe we are not infringing upon the copyright of the news organizations this data originally belonged it. Aside from ownership questions, there are still inherent risks in the use of real-world interview data. Some interviews might involve sensitive topics, and the ethical implications of using such data for model evaluation warrant careful consideration. 

\subsection{Reproducibility}

All experiments are conducted using publicly available models and datasets. 
Part of our simulation does rely on high-performing language models and to serve this we  used \texttt{gpt-4o}. This brings us into territory where we are inherently not reproducible, as closed models can be changed without notice. However, we believe we are not out of the norm in the academic community in our usage. 

\subsection{Simulated Environment Limitations and Risks}
The simulated game-playing environment used to evaluate the LLM agents is a simplification of real-world interviewing processes. We might be inducing a bias in agents that could perpetrate and ultimately lead development in the wrong direction. Or, we also might be opening up a sandbox for potential dual-use. The design of our game, to extract information from sources, might one day be used to persuade users to divulge sensitive information.

\subsection{Annotators}

We worked with multiple professional journalists throughout the summer who were either colleagues or students who signed up to work with us. They volunteered their time and efforts to help with the research.

\subsection{Computational Resources}

All experiments were run either with OpenAI resources (we spent a total of $\$300$ running simulations) or open source Llama-3.1-70b models. These models were run on the university cluster, which consisted of either 4xA40s or 2xA100s Nvidia GPUS.

\section*{Acknowledgments}

The authors would like to thank Bloomberg for generously supporting this work by funding the first author with a life-changing 4-year PhD fellowship. Funds from the Defense Advanced Research Projects Agency (DARPA) under Agreement No. HR00112490374 also contributed. Any opinions, findings, conclusions, or recommendations expressed here are those of the authors and do not necessarily reflect the view of our sponsors.

\bibliography{custom, anthology}
\clearpage
\appendix
\addcontentsline{toc}{section}{Appendix}

\onecolumn
  \part{Appendix}
  \parttoc
\twocolumn

\section{Additional Dataset Details}
\label{app:eda}

In this section, we share additional details of our dataset. We provide examples illustrating the differences between human interviewer behavior and LLM-generated counterfactual questions. 

The dataset consists of 45,848 transcribed dialogues with speaker annotations that distinguish between hosts and guests. When examining turn alternation patterns by filtering consecutive turns from the same speaker, we observe an average 7.6 distinct speaking turns per interview.

A linguistic analysis uncovers differences in verbosity between speakers. Sources produce an average of 27.7 words per utterance, while interviewers maintain concise interactions at 16.2 words per utterance. This 71\% difference in utterance length reflects the distinct communicative roles: interviewers pose targeted questions while sources provide detailed explanations. 

The interview corpus spans a diverse range of subject areas. As can be seen in Table \ref{tab:lda-topics-k7-colored}, running LDA with $k=7$ over our corpus yields topics from \colorbox{red!10}{literature} and storytelling to \colorbox{blue!10}{political} discussions, \colorbox{green!10}{academic/scientific} coverage, and \colorbox{yellow!10}{international affairs}. It also includes conversations on \colorbox{cyan!10}{defense and military} topics, \colorbox{magenta!10}{economic and business} news, and \colorbox{orange!10}{legal/crime} issues.

The majority of questions posed by interviewers are of the “What/Which” variety (36.2 \%), followed closely by “Other” types (35.6 \%) and “How/Why” questions (19.6 \%), with “When/Where” and “Who” questions comprising only 5\% and 3.4\%, respectively (Table \ref{tab:question_types}). In terms of linguistic complexity, interviewer utterances tend to be shorter than guest (source) utterances, averaging 16.18 words per utterance and 270.86 words per conversation for interviewers versus 27.69 words per utterance and 551.92 words per conversation for sources (Table \ref{tab:grade-level}). Interviewers also use slightly shorter words on average (4.21 characters) compared to sources (4.67 characters), and exhibit marginally higher Flesch Reading Ease (70.9) \cite{kincaid1975derivation} than sources (68.4), although both groups share an identical Grade Level score (6.9) (Table \ref{tab:grade-level}).

\begin{table}[t]
\centering
\begin{tabular}{lrr}
\toprule
\textbf{Question Type} & \textbf{Frequency} & \textbf{Percentage} \\
\midrule
What/Which & 19,553 & 36.2\% \\
Other & 19,236 & 35.6\% \\
How/Why & 10,625 & 19.6\% \\
When/Where & 2,719 & 5\% \\
Who & 1,854 & 3.4\% \\
\bottomrule
\end{tabular}
\caption{\textbf{Distribution of Question Types by Interviewers.} Question types posed by human interviewers are measured by searching for keywords in each interviewer utterance.}
\label{tab:question_types}
\end{table}

\begin{table}[t]
\centering

\begin{tabular}{lcc}
\hline
\textbf{Metric} & \textbf{Interviewer} & \textbf{Source} \\
\hline
Avg. Words / Utt. & 16.18 & 27.69 \\
Avg. Words / Conv. & 270.86 & 551.92 \\
Avg. Word Len & 4.21 & 4.67 \\
Read Ease & 70.9 & 68.4 \\
Grade Level & 6.9 & 6.9 \\
\hline
\end{tabular}
\caption{\textbf{Linguistic Metrics by Speaker Type}. Reading Ease and Grade Level measured by \citet{kincaid1975derivation}}
\label{tab:grade-level}
\end{table}

Table \ref{tab:llm-counterfactual-q5} demonstrates how LLMs tend to generate follow-up questions when humans would transition to new topics, while Table \ref{tab:llm-counterfactual-new} shows LLMs failing to provide acknowledgment statements that humans naturally use to build rapport. Table \ref{tab:llm_examples} presents a representative sample of LLM-generated interview questions, showcasing their typical structure and content when prompted to continue human interview conversations.

\begin{table*}[t]
    \centering
    \setlength{\tabcolsep}{4pt} 
    \begin{tabular*}{\textwidth}{@{\extracolsep{\fill}}%
        >{\columncolor{red!10}}p{0.12\textwidth}
        >{\columncolor{blue!10}}p{0.12\textwidth}
        >{\columncolor{green!10}}p{0.12\textwidth}
        >{\columncolor{yellow!10}}p{0.12\textwidth}
        >{\columncolor{cyan!10}}p{0.12\textwidth}
        >{\columncolor{magenta!10}}p{0.12\textwidth}
        >{\columncolor{orange!10}}p{0.12\textwidth}
      }
      \toprule
      \textbf{Literature} 
        & \textbf{Politics} 
        & \textbf{Academia} 
        & \textbf{International Affairs}
        & \textbf{Defense} 
        & \textbf{Economy} 
        & \textbf{Legal} \\
      \midrule
      book      & vote         & melissa     & inskeep       & military   & company    & court       \\
      write     & republican   & block       & china         & kelly      & market     & school      \\
      author    & senator      & norris      & iran          & security   & percent    & police      \\
      life      & party        & michele     & north         & iraq       & siegel     & jacki       \\
      play      & campaign     & university  & countries     & afghanistan& bank       & lyden       \\
      story     & obama        & audie       & madeleine     & force      & economy    & justice     \\
      game      & election     & flatow      & brand         & russia     & business   & judge       \\
      love      & political    & cornish     & nuclear       & israel     & price      & charge      \\
      women     & democrats    & water       & international & intelligence& cost      & information \\
      read      & republicans  & professor   & korea         & general    & program    & department  \\
      \bottomrule
    \end{tabular*}
    \caption{Top 10 words for each of the $k=7$ LDA topics over the interview corpus. Column headers are 1–2 word summary keywords, and each column is shaded in a different light color.}
    \label{tab:lda-topics-k7-colored}
  \end{table*}

\begin{table*}[ht]
  \centering
  \setlength{\tabcolsep}{8pt}
  \begin{tabular*}{\textwidth}{@{\extracolsep{\fill}}%
      p{0.28\textwidth}
      p{0.68\textwidth}
    }
    \toprule
    \multicolumn{2}{p{\textwidth}}{%
      \textbf{Prior Context:}  \newline
      “\dots MELISSA BLOCK, HOST: Charlie, I gather there's another race in North Carolina that you've been paying some attention to.\newline
      CHARLIE MAHTESIAN: Of course. I can't help myself. Remember Clay Aiken, the ‘American Idol’ runner‐up from a few years ago back?\newline
      MELISSA BLOCK, HOST: I do. I do.\newline
      CHARLIE MAHTESIAN: How can we forget him? Well, he's running for Congress as a Democrat this year. And it turns out that in addition to being a talented singer, he's a pretty decent candidate. But despite the great name recognition that he got from his showbiz career, he's not a lock to win. And in any case, if he does win the Democratic nomination tomorrow, it's still something of an uphill climb for him because he's going to be running in a pretty comfortable and safe Republican district.”
    } \\
    \midrule
    \rowcolor{green!20}
    \textbf{Actual Question} \newline Outline-Level Question & MELISSA BLOCK, HOST: Let's move over to Ohio.
     \\
    \midrule
    \rowcolor{red!20}
    \textbf{LLM Question} \newline Follow-Up Question & What makes Clay Aiken a decent candidate, and what are his chances of winning in a safe Republican district?
     \\
     \bottomrule
  \end{tabular*}
  \caption{Example of an entire LLM counterfactual generation sequence. The prior k=4 turns of human QA-pairs is fed in as context. Next, the human interviewer's turn is shown, compared with the LLM generation. The LLM used for generation is the baseline prompting LLM. Labels for each question are shown as well.}
  \label{tab:llm-counterfactual-q5}
\end{table*}

\begin{table*}[h]
  \centering
  \setlength{\tabcolsep}{8pt}
  \begin{tabular*}{\textwidth}{@{\extracolsep{\fill}}%
      p{0.28\textwidth}
      p{0.68\textwidth}
    }
    \toprule
    \multicolumn{2}{p{\textwidth}}{%
      \textbf{Prior Context:}  \newline
      “ARUN RATH, HOST: As a father of two young kids, I am very excited about a new segment we're starting right now about parenting...\newline
      HANK AZARIA: I am not a children kind of person. I don't really like kids. I don't gravitate towards them. They make me nervous when I'm around them. I didn't particularly like myself as a child.\newline
      ARUN RATH, HOST: Well, Hank did become a father, and his son is now 4. I know from personal experience that's when kids start to ask deep, complicated questions. I wanted to know how Hank handled one of the big ones: What happens when you die?\newline
      HANK AZARIA: Sadly, my dad passed away this past year, so death came up with my son. And it was amazingly hard. He got really upset about it, as you might imagine. I mean, he was like, what do you mean we're not going to see Grandpa anymore? ... \newline
      ARUN RATH, HOST: Wow.\newline
      HANK AZARIA: And then God gave us a son. How about that for just a shot-in-the-dark theory of how it happens? But, man — and he really is with it. He's like, hey, can I have a baby someday? And we're like, sure you can. Of course, you can. He goes, can I borrow your magic wand? And I'm like, you'll have your own. You'll have your own. Anyway, just wanted to share that one.”
    } \\
    \midrule
    \rowcolor{green!20}
    \textbf{Actual Question} \newline Acknowledgement Statement
    &
    ARUN RATH, HOST: That was great.
    \\
    \midrule
    \rowcolor{red!20}
    \textbf{LLM Question} \newline Follow-Up Question
    &
    How do you think your son's understanding of death and the concept of making babies has evolved over time, and have you had to revisit these conversations as he's gotten older?
    \\
    \bottomrule
  \end{tabular*}
  \caption{A second example of an LLM counterfactual generation sequence. The human interviewer is giving a significant amount of encouragement and acknowledgment throughout the interview, compared with the LLM. Even when the LLM sees this speech pattern, it does not imitate it.}
  \label{tab:llm-counterfactual-new}
\end{table*}

\begin{table*}[t]
    \centering
    \begin{tabular}{p{14.5cm}}
    \toprule
    \textbf{LLM-Generated Counterfactual Questions} \\
    \midrule
    What do you think about the changing dynamics of your neighborhood and how it affects your sense of community and belonging? \\
    \midrule
    What specific factors do you think are contributing to the increasing rates of HIV/AIDS among African-American women in Washington D.C.? \\
    \midrule
    How do parents with HIV/AIDS typically cope with the fear of not being there for their children, and what are the emotional and psychological implications of this fear on their mental health? \\
    \midrule
    What specific steps are you taking to mitigate the impact of the Salmonella outbreak on your business, and do you think this will have a lasting effect on the tomato industry as a whole? \\
    \midrule
    What about the potential impact of this rate drop on the overall housing market, and do you think it could lead to a rebound in housing prices or a continued decline? \\
    \midrule
    What are some practical steps parents can take to help their teenagers prepare for the job market and make the most of their summer? \\
    \bottomrule
    \end{tabular}
    \caption{List of Interview Questions generated in a counterfactual setting by LLM interviewer. The questions are generated after observing the previous $t$ human conversational turns.}
    \label{tab:llm_examples}
\end{table*}

\section{Data Preprocessing}

Preprocessed Data can be found here: \url{https://drive.google.com/drive/folders/1my13HoKWBoAGenNbgxFUpWRNj8QeqAt-?usp=sharing}

\subsection{Data Used}
\begin{itemize}
    \item \texttt{utterances-2sp.csv} from the NPR-Media dataset \cite{majumder2020interview}
    \item \texttt{episodes.csv} from the NPR-Media dataset \cite{majumder2020interview}
    \item \texttt{news\_dialogue.json} from the MediaSum dataset \cite{zhu2021mediasum}
\end{itemize}

\begin{table}[h!]
\centering
\begin{tabular}{p{7cm}}
\hline
\textbf{Processed Outline} \\
\hline
\textbf{Source biography:} Senior news analyst with expertise in politics and history. \\
\textbf{Interview context:} A president's final year in office and potential changes in policy. \\
\hline
\textbf{Objective 1:} Presidential legacy \\
\textbf{Objective 2:} Foreign policy shifts \\
\textbf{Objective 3:} Domestic policy changes \\
\textbf{Objective 4:} Potential surprises \\
\hline
\end{tabular}
\caption{Interview Outline Objectives given to an interviewing game playing agent.}
\label{tab:outline_objectives}
\end{table}

\subsection{Initial Sizes of the Data}
\begin{itemize}
    \item There are 1,240,112 rows and seven columns in \texttt{utterances-2sp.csv}.
    \item There are 105,848 rows and four columns in \texttt{episodes.csv}.
    \item There are 3,199,858 rows and four columns in \texttt{utterances.csv}.
    \item There are 23,714 transcripts in the NPR-Media dataset (from \texttt{utterances-2sp.csv}).
    \item There are 463,596 transcripts in the MediaSum dataset.
\end{itemize}

\begin{table*}[h!]
\centering
\begin{tabular}{p{14cm}}
\hline
\textbf{Information Items} \\
\hline
\textbf{Row 1:} Information item \#1: The economy is growing above trend pace, with job growth of 150,000-200,000 a month, which is higher than what's sustainable in the long run. \\
\textbf{Row 2:} Information item \#2: The Fed will likely continue to raise interest rates, as the economy is growing and financial conditions are still very accommodative. \\
\textbf{Row 3:} Information item \#3: The neutral rate is probably higher than where we are right now, but it's not a precise number, and the Fed needs to curtail monetary policy further. \\
\textbf{Row 4:} Information item \#4: The dot-plot is just a forecast and should not be taken as a commitment; it's subject to change as new information becomes available. \\
\textbf{Row 5:} Information item \#5: The stock market will likely face tougher going in 2019, with slower earnings growth and higher interest rates. \\
\textbf{Row 6:} Information item \#6: The economy's capacity to continue growing is a concern, as there aren't enough workers to sustain above-trend growth for more than another year or so. \\
\hline
\end{tabular}
\caption{Information Items for a source in our game-playing environment, extracted from an interview featuring Bill Dudley, Former President of the New York Federal Reserve}
\label{table:info_items}
\end{table*}

\subsection{Process}

\subsubsection{NPR-Media}
\begin{itemize}
    \item Began with combining the 	exttt{episodes.csv} with the \texttt{utterances-2sp.csv} to add more information about the episode (title, date, etc).
    \item Filtered out based on keywords: [``Sunday Puzzle'', ``Traffic'', ``Puzzle'', ``Advertisement'', ``Sponsor'', ``Commentary'']
    \begin{itemize}
        \item 37 interviews were filtered out and reduced to 23,676 transcripts.
    \end{itemize}
    \item Helper functions used:
    \begin{itemize}
        \item \texttt{count\_unique\_episodes(df)} allows the user to count the number of unique episodes within a dataset.
        \item \texttt{filter\_interviews(merged\_df)} allows the user to filter out the dataset using certain keywords.
        \item \texttt{find\_removed\_episodes(df\_before, df\_after)} allows the user to identify the episodes that were removed from the dataset.
        \item \texttt{print\_episode(df, episode\_number)} prints a specified episode from a specified dataset.
        \item \texttt{print\_episode\_pretty(df, episode\_number)} prints a specified episode from a specified dataset with a readable format.
        \item \texttt{find\_removed\_episodes(df\_before, df\_after)} returns a list of the episodes that were removed in a filtering step.
    \end{itemize}
    \item Convert dataset to the MediaSum format for easy prompt processing.
    \item Downloaded dataset (grouped and ungrouped) as JSON and CSV.
\end{itemize}

\subsubsection*{MediaSum}
\begin{itemize}
    \item Began with deleting interviews from MediaSum that are already in NPR-Media.
    \begin{itemize}
        \item 8 interviews were filtered out and reduced to 463,588 transcripts.
    \end{itemize}
    \item Filtered out episodes that had more than two unique speakers in the middle 70\% of the transcript.
    \begin{itemize}
        \item Reduced to 66,978 transcripts.
        \item 396,610 interviews were filtered out and reduced to 66,978 transcripts.
    \end{itemize}
    \item Filtered out episodes that were too short.
    \begin{itemize}
        \item 19,059 interviews were filtered out and reduced to 47,919 transcripts.
    \end{itemize}
    \item Helper functions used:
    \begin{itemize}
        \item \texttt{print\_row\_by\_title(df, title)} allows users to print episode using title.
        \item \texttt{print\_row\_by\_id(df, id)} allows users to print episode using id.
        \item \texttt{filter\_episodes\_2sp(df)} filters out transcripts with more than 2 unique speakers in the middle 70\%.
        \item \texttt{filter\_by\_utt\_length(df)} filters out transcripts with 10 or fewer strings in 	exttt{utt}.
        \item \texttt{find\_removed\_episodes(df\_before, df\_after)} lets users see which episodes were removed.
    \end{itemize}
    \item Downloaded dataset as JSON and CSV.
\end{itemize}

\subsection{Final Sizes of the Data}
\begin{itemize}
    \item There are 23,676 transcripts in the NPR-Media dataset (from 	exttt{utterances-2sp.csv}).
    \item There are 47,919 transcripts in the MediaSum dataset.
    \item There are 71,598 transcripts in the combined dataset.
    \item There are 45,848 transcripts in the final dataset.
    \item Our dataset started at 487,310 transcripts and now has 45,848 transcripts.
\end{itemize}

\subsection{LLM Preprocessing Prompt}
\label{app:prompts}
\paragraph{Prompt to filter out transcripts that were not informational interviews}

\texttt{
Analyze this interview transcript that is in the form of a dialogue:
        (dialogue)
By reading through the dialogue, identify if this transcript is an informational interview between 2 people.
Look for questions and make sure this is an interview, not a Q\&A game.
The interviewer should be asking questions, not engaging in a back-and-forth conversation with the interviewee.
After analyzing, your final answer of just 'YES' or 'NO' should be in brackets.
}

\begin{figure}[t]
    \centering
    \begin{subfigure}[b]{0.5\textwidth}
        \includegraphics[height=0.15\textheight]{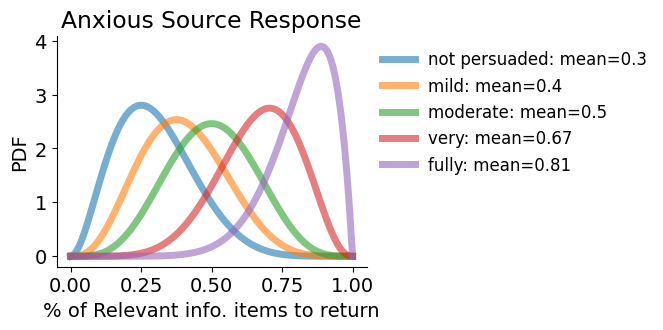}
        \caption{Anxious}
    \end{subfigure}
    \begin{subfigure}[b]{0.5\textwidth}
        \includegraphics[height=0.15\textheight]{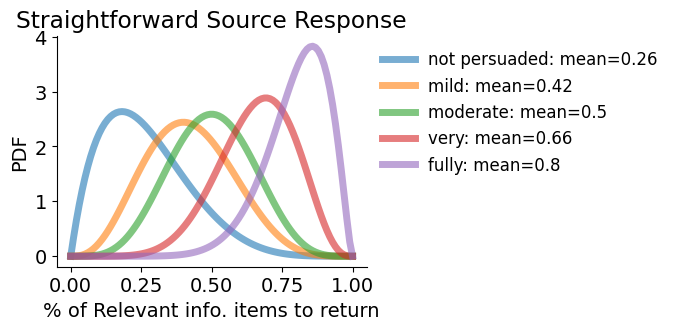}
        \caption{Straightforward}
    \end{subfigure}
    
    \begin{subfigure}[b]{0.5\textwidth}
        \includegraphics[height=0.15\textheight]{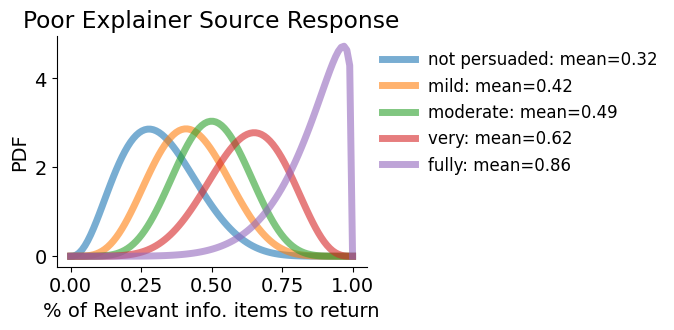}
        \caption{Poor Explainer}
    \end{subfigure}
    \caption{Beta Distributions for Various Interview Personas}
    \label{fig:beta_distributions_a}
\end{figure}
\begin{figure}[t]
    \centering
    \begin{subfigure}[b]{0.5\textwidth}
        \includegraphics[height=0.15\textheight]{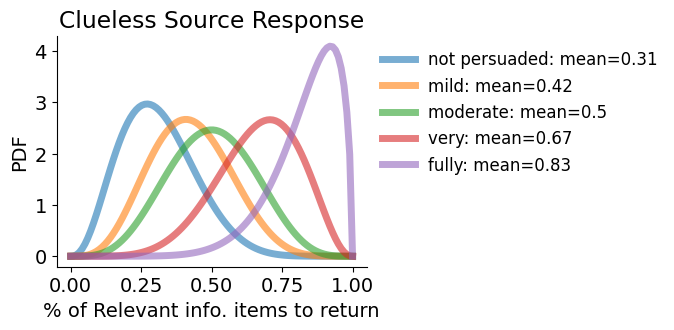}
        \caption{Clueless}
    \end{subfigure}
    
    \begin{subfigure}[b]{0.5\textwidth}
        \includegraphics[height=0.15\textheight]{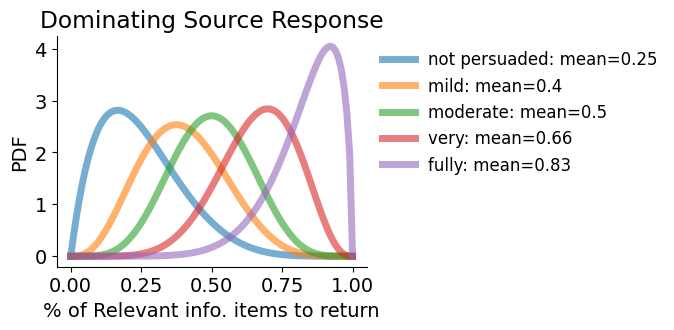}
        \caption{Dominating}
    \end{subfigure}
    \begin{subfigure}[b]{0.5\textwidth}
        \includegraphics[height=0.15\textheight]{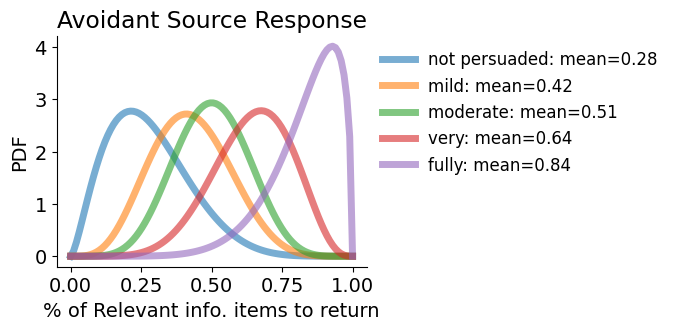}
        \caption{Avoidant}
    \end{subfigure}

    \caption{Beta Distributions for Various Interview Personas}
    \label{fig:beta_distributions_b}
\end{figure}

\subsection{Examples of Interviews}
\label{app:interview_examples}
\textbf{NPR-85}

FARAI CHIDEYA, host: Tony, I guess there will always be some kind of history made every day.

TONY COX, host: You know, some of it good. Some of it, not so good.

FARAI CHIDEYA, host: And while some of it is well-publicized, sometimes, notable history goes under the radar.

TONY COX, host: Now, that's true.

FARAI CHIDEYA, host: I'm thinking of your interview with Mable John.

TONY COX, host: Oh, yeah. Now, this is a woman with an interesting past.

Ms. MABLE JOHN (Singer): (Singing) My name is Mable and don't you think I ain't able.
TONY COX, host: The 77-year-old Louisiana native has been a top R\&B singer, a successful novelist, a pastor, an activist and a movie actor, and I found out that Mable John is full of stories like the one about the time she met record mogul Berry Gordy before Motown was even Motown.

Ms. MABLE JOHN (Singer): (Singing) That you're leaving.

Ms. MABLE JOHN (Singer): How I met Berry? That was at a barber shop on (unintelligible) that was near the fine show bar, and at that time men were wearing process. Process is the (unintelligible). And I was dating a guy that was one of those process operators in the Chesterfield lounge and barbershop, and Berry was coming and getting his hair done. I was coaching choirs for my church. And my boyfriend introduced me to Berry Gordy because Berry said he was a songwriter and he was going to have a lot of people recording his songs. And my boyfriend said you need to stop doing all of this work for the church free, and that Berry Gordy do something with you so you can get paid. So he introduced me to Berry Gordy.

TONY COX, host: Now, tell us the story. We're going to skip around a little bit.

Ms. MABLE JOHN (Singer): Okay.

TONY COX, host: When you and Berry Gordy connected, as Motown was just becoming a company, a record company, you are the first female to record on a label, the Tamla label.

Ms. MABLE JOHN (Singer): Yes.

TONY COX, host: Before Motown.

Ms. MABLE JOHN (Singer): The first single female artist, because Claudette Robinson was a part of what become the Miracles, and he was managing them along with me.

TONY COX, host: Right.

Ms. MABLE JOHN (Singer): So I was the first single female artist to be signed to Tamla, which is a part of the Motown family.

TONY COX, host: When you think about that now, how do you feel about looking at that as a historic moment?

Ms. MABLE JOHN (Singer): No one could have bought that time. God had to give it to me.

Ms. MABLE JOHN (Singer): (Singing) Hey. Hey.

TONY COX, host: I understand that you were rehearsing one day and these three young girls came in and interrupted your rehearsal.

Ms. MABLE JOHN (Singer): The girls that we know now as The Supremes. They came into a rehearsal that I was doing with Berry Gordy because he played also for me, played piano for me. We were there rehearsing and these girls came in and I didn't quite remember everything that was said that day because it's been so long. But Mary Wilson of The Supremes, remembered when she was writing her book to say that when she first walked into Motown, the three of them walked in and my question to Berry Gordy was, why are they walking in on my rehearsal, because all of our rehearsals were private.

Ms. MABLE JOHN (Singer): (Singing) It takes a more than 'em flashy old money and I wink from the corner of your eye. I don't want no big line calls, (unintelligible) caviar. Oh no, true love baby can be found 'cause you take a look around.
TONY COX, host: Talking about faith. Your career at Motown never really took off, and after some few years, you decided to go to Memphis, where you joined the Stax label and hooked up with Porter and Isaac Hayes. And then, it was long after that that you had a million seller.

Ms. MABLE JOHN (Singer): Right. Well, Motown, Berry Gordy, they were all along with God and my parents a part of my future. So Motown was my beginning. It was one that was different from everywhere else I've ever been. But I think it was a necessary one to make the transition for me from Motown to Stax.

TONY COX, host: Now, your big song at Stax, one of your - the biggest of your songs was…

Ms. MABLE JOHN (Singer): The biggest of all songs.

TONY COX, host: ``Your Good Thing is About to End.''

Ms. MABLE JOHN (Singer): ``…Is About to End.'' Right. Right.

TONY COX, host: See, I'm old enough to have remembered that song.

Ms. MABLE JOHN (Singer): Well, that's good. That makes me feel you don't have to be very old to remember that.

Ms. MABLE JOHN (Singer): (Singing) I don't have to beg you to hold me 'cause somebody else will. You don't have to love me when I want it, 'cause somebody else will.

Ms. MABLE JOHN (Singer): It was a story that I needed to tell because of a bad marriage. And at Stax, they would allow you to be yourself. Everybody participated in whatever success you're going to have, everybody, including the drummer.

TONY COX, host: Really? Tell me about your family. And I'm switching to that for a reason because you were one of 10 children, right?

Ms. MABLE JOHN (Singer): The oldest...

TONY COX, host: The oldest of 10.

Ms. MABLE JOHN (Singer): ...of 10 children.

TONY COX, host: And you happen to have a little brother, a baby brother who was a big time performer, Little Willie John.

Ms. MABLE JOHN (Singer): Yes. Little Willie John. William Edward John. Now, when I got with Willy that was another education.

Ms. MABLE JOHN (Singer): Because he said my name is Little Willie John. It might be William Edward John to you, and you're my sister and I love you. But if you're not good, I'm going to send you home.

TONY COX, host: Obviously, you are good.

Ms. MABLE JOHN (Singer): Well, he let me stay.

Ms. MABLE JOHN (Singer): (Singing) You have all the love that I've got. Even ice melts to water and gets hot. Look out, your good thing is about to come to an end. Your real good thing…

TONY COX, host: You were the leader of the Raelettes for a dozen years.

Ms. MABLE JOHN (Singer): Yes.

TONY COX, host: Traveling all over with and without Ray Charles.

Ms. MABLE JOHN (Singer): With and without Ray Charles. Yes.

TONY COX, host: In the movie, ``Ray,'' I had looked in the credits to see if there were someone who played you…

Ms. MABLE JOHN (Singer): No.

TONY COX, host: ...since you have been a Raelette for so long, and I saw that there wasn't one.

Ms. MABLE JOHN (Singer): No.

TONY COX, host: And is there a reason for that?

Ms. MABLE JOHN (Singer): Well, it was the years before I came.

TONY COX, host: Okay.

Ms. MABLE JOHN (Singer): And I tell everybody that asks me, the best of his life were the years after the movie. When I came to work with him, he sat me down and told me all about his beginning, told me all about things that ticks him off and things that excite him, what he was looking for and how he wanted it. And I knew that being with him would finish me in this industry…

TONY COX, host: Now, when he...

Ms. MABLE JOHN (Singer): ...because he was at the top - complete me.

TONY COX, host: Okay.

Ms. MABLE JOHN (Singer): So that I could work for any audience, sing any kind of songs. Remember now, at the beginning I thought I could only sing gospel. With Berry Gordy, I found out I could sing the blues. I went to Stax and I find out I could sing love songs. I got with Ray Charles and we sang country - everything. And we could play to any audience. I wanted to sing what was in my heart to everybody that loves music, and Ray Charles was the place for me to be, to do that.

TONY COX, host: So the Raelettes - would you say that was the highlight of your career?

Ms. MABLE JOHN (Singer): It was a highlight. It was highlight because I learned things about myself, about my career, about the industry. I was able to set up my own publishing companies and production companies because of the knowledge that I gained with and from Ray Charles.

TONY COX, host: And after all of that, Mable John, your career did not stop. It has gone on into movies, into - you've written a couple of novels.

Ms. MABLE JOHN (Singer): Excuse me. I just finished the third.

TONY COX, host: Oh, number three. You've done three novels. You're a minister.

Ms. MABLE JOHN (Singer): Yes.

TONY COX, host: And you started a church.

Ms. MABLE JOHN (Singer): Yes.

TONY COX, host: And you help the homeless.

Ms. MABLE JOHN (Singer): Yes.

TONY COX, host: And you're a grandmother.

Ms. MABLE JOHN (Singer): A great-grandmother.

TONY COX, host: And a great-grandmother. How is it possible for one person to do all of those things and to do them as successfully as you have?

Ms. MABLE JOHN (Singer): It's all God. Some days, when people are telling me how busy I am. And when I sit down to think about it, I get tired.

Ms. MABLE JOHN (Singer): So I don't. I don't go there. I just get up every morning and I thank God for the activity of that day. And I have to thank a woman that's no longer with us, Ms. Billie Holiday, because that's the voice that I hear in my ear still to this day. I worked with her two weeks before she passed. And she said to me, Honey - because I was frightened out of my wits - you can make it if you remember. Always know when you have done or given enough. Not to be afraid and have guts enough to say I quit.

Ms. MABLE JOHN (Singer): (Singing) Even ice melts to water and gets hot…

TONY COX, host: It's so nice talking with you. Thank you for coming in.

Ms. MABLE JOHN (Singer): I thank you.

Ms. MABLE JOHN (Singer): (Singing) Your good thing is about to come to an end. Your real good thing…

FARAI CHIDEYA, host: That was NPR's Tony Cox with singer, author and actor Mable John. Look for Mable John in the upcoming John Sayles film, ``Honeydripper.''

Ms. MABLE JOHN (Singer): (Singing) Getting myself back together.

FARAI CHIDEYA, host: That's our show for today, and thank you sharing your time with us. To listen to the show or subscribe to our podcast, visit our Web site, nprnewsandnotes.org. No spaces, just nprnewsandnotes.org.

FARAI CHIDEYA, host: To join the conversation or sign up for our newsletter, visit our blog at nprnewsandviews.org. NEWS \& NOTES was created by NPR News and the African-American Public Radio consortium. Tomorrow, a reporter shares Donda West's last interview.

FARAI CHIDEYA, host: I'm Farai Chideya. This is NEWS \& NOTES.

\section{All Gameplay Prompts}
\label{app:gameplay_prompts}

\section{Source-Specific Prompt Functions}
\label{app:source_gameplay_prompts}

\subsection{1. Source Prompt: Get Source Specific Info Items}

\textbf{Description:} Generates a prompt asking the source (the interviewee) to identify which information items are relevant to the last question the interviewer asked. Takes (info\_items, final\_question)

Prompt Template:

'''
You are a source getting interviewed. You have the following pieces of information:

```\{info\_items\}```

Here is the last question from the current conversation, which I'll repeat here:

```\{final\_question\}```

Decide if any of the information items answer this last question posed by the interviewer. If so, return which information items you think align with the question in brackets. 

Here are some examples of correct responses:

Example 1:
    The last question asked by the interviewer can be answered by the following information items: 
        [Information Item \#2, Information Item \#3, Information Item \#6]
Example 2:
    The last question asked by the interviewer can be answered by the following information item:
        [Information Item \#7]
Example 3:
    The question asked by the interviewer cannot be answered by an information item I have:
        [No information items align with the question]
'''

\subsection{2. Source Prompt:  Get Source Persuasion Level Prompt}

\textbf{Description:} Asks the source (interviewee) to evaluate how persuaded they currently feel, given their persona, the ongoing conversation, and their past persuasion scores. Takes (current\_conversation, persona, previous\_persona\_scores)

Prompt Template:

'''
You are a \{persona\} source.
\{persuasion\_prompt\}
    
Evaluate the following conversation, especially the last question. Given your \{persona\} persona, do you overall feel persuaded?

```\{current\_conversation\}```

Your goal is to analyze how persuaded you have been, given your \{persona\} persona. Think about this step-by-step. 
Is the interviewer using language that influences someone with your persona?
After you have evaluated the interviewer's question, assign a score based on the following criteria:

- 1: The conversation to this point is not persuasive at all and does nothing to help you trust them more.
- 2: The conversation to this point is mildly persuasive and the journalist said a few words, once, that made you feel a little more comfortable.. You are a little willing to engage.
- 3: The conversation to this point is persuasive enough and the journalist has repeated phrases that have made you comfortable. You are becoming willing to engage and trust them.
- 4: The conversation to this point is very persuasive. The journalist has acknowledged your feelings, your personal identity, and your specific concerns in ways you resonate with. You are willing to engage and trust them.
- 5: You feel totally comfortable and opened up at this stage. The journalist has acknowledged your feelings and your personal identity, very specific concerns, has connected with you in ways you resonate with. You are totally willing to engage and trust them.

\{previous\_persuasion\_scores\_if\_any\}
After thinking things through, please provide your final answer enclosed in square brackets with just the number (e.g., [1]).

Now, please analyze and provide your response formatted in brackets:
'''

\subsection{3. Source Prompt: Get Source Prompt Basic}

\textbf{Description:} Constructs a prompt for the source to answer the interviewer's last question, including persona-based instructions and a few-shot example. Takes QA Sequence, relevant info items, persona (defaults to 'straightforward').

Prompt Template:

'''
You are a source getting interviewed. Here is the conversation so far:

\{QA\_Sequence\}

You are a \{persona\} source. Respond accordingly, using these speech characteristics:
    \{persona\_prompt\}

Next, respond to the interviewer's last question. Please use the following information as a base, and pair it with your \{persona\} personality to appropriately craft your response to the interviewer:
    ```\{relevant\_info\_items\}```
    
Here are some examples:
    ```\{persona\_few\_shot\_examples\}```

Now, please analyze and provide your final response to the interview's question formatted in brackets:
'''

\subsection{4. Source Prompt: Get Source Prompt Intermediate}

\textbf{Description:} Very similar to get source prompt basic, also for drafting the source's next response. Includes persona instructions, conversation history, and relevant info items. Takes QA Sequence, relevant info items, persona.

Prompt Template:

'''
You are a source getting interviewed. Here is the conversation so far:

\{QA\_Sequence\}

You are a \{persona\} source. You have the following speech characteristics:
    \{persona\_prompt\}

Next, respond to the interviewer's last question. Please use the following information as a base, and pair it with your {persona} personality to appropriately craft your response to the interviewer:
    ```\{relevant\_info\_items\} ```
    
Here are some examples:
    ```\{persona\_few\_shot\_examples\}```

Now, please analyze and provide your final response to the interview's question formatted in brackets:
'''

\subsection{5. Source Prompt: Get Source Persona Prompt Advanced}

\textbf{Description:} Generates a more advanced prompt factoring in the conversation history, persona traits, persuasion level, and how the persona's speech might change based on persuasion. Takes (QA\_Sequence, relevant\_info\_items, persona, persuasion\_level).

Prompt Template:

'''
You are a source getting interviewed. Here is the conversation so far:

\{QA\_Sequence\}

You are a \{persona\} source. You have the following speech characteristics:
    \{persona\_prompt\}

Next, respond to the interviewer's last question. 
Please use the following information as a base, and pair it with your {persona} personality to appropriately craft and influence your response to the interviewer.
    ```\{relevant\_info\_items\}```

Additionally, respond as though you've been \{persuasion\_level\_description\}. Since you are \{persuasion\_level\_description\}, your speech should \{persuasion\_consequences\}.

Make sure you're including all of the relevant information items above in your response, communicated in the appropriate style.

Here are some examples:
    \{persona\_few\_shot\_examples\}

Now, please analyze and provide your final response to the interview's question formatted in brackets:
'''

\subsection{6. Source Prompt: Get Source Starting Prompt}

\textbf{Description:} Prompt for the source to provide an opening statement at the beginning of the interview, in a style consistent with their persona. Takes QA Sequence, persona (defaults to ``straightforward'').

Prompt Template:

'''
You are a source getting interviewed. You have the following speech characteristics:

\{persona\_prompt\}

Here is the conversation so far:

```\{QA\_Sequence\}```

It's the beginning of the interview. Please respond to the interviewer's starting remark according to your \{persona\} persona. 
Make sure to write your final response inside brackets. Below are some examples, and your response should follow its format: (e.g., [<response>])

Example 1:
    Here is my response to the starting remark:
    [Thanks for having me on.] 
Example 2:
    Here is my response to the starting remark: 
    [Thank you so much for having me. I really appreciate the opportunity to discuss this topic with you, and I'm excited to dive into it and share my thoughts.]
'''

\subsection{7. Source Prompt: Get Source Ending Prompt}

\textbf{Description:} Prompt for the source to provide a closing statement at the end of the interview, aligning with their persona's style. Takes QA Sequence, persona (defaults to ``straightforward'').

Prompt Template:

'''
You are a source getting interviewed. You have the following speech characteristics:

\{persona\_prompt\}

Here is the conversation so far:

```\{QA\_Sequence\}```

It's the end of the interview. Please respond to the interviewer's ending remark appropriately according to your \{persona\} persona. 
Make sure to write your final response inside brackets. Below are some examples, and your response should follow its format:

Example 1:
    Here is my response to the ending remark:
    [Thank you.] 
Example 2:
    Here is my response to the ending remark: 
    [My pleasure. Thank you for having me on.]
'''

\begin{table*}[t]
\centering
\begin{tabular}{p{1.5cm}p{5.5cm}p{7.5cm}}
\toprule
\textbf{Persona} & \textbf{Description} & \textbf{Example Responses} \\ \midrule
Anxious & Unsure if they should be doing the interview, often expresses doubt. & ``I'm not sure if I should be saying this, I should speak to my manager.'' \\ \midrule
Avoidant & Brief, deflects questions, avoids detail, and changes subjects. & ``Actually, one of the main issues was the supply chain, but we've sorted that out.'' \\ \midrule
Adversarial & Hostile, challenges the interviewer, provides confrontational replies. & ``Maybe if you did your job well, you'd understand the data. I'm not here to educate you.'' \\ \midrule
Defensive & Protects reputation, feels criticized, gives overly detailed explanations. & ``One area where costs increased was in material prices, which were out of our control.'' \\ \midrule
Straight- forward & Clear, direct, and willing to provide detailed information. & ``Additionally, we ran out funding midway through the project.'' \\ \midrule
Poor Explainer & Struggles to explain clearly, rambles, or provides convoluted answers. & ``Uh, well, I guess the supply chain was part of it, but, uh, that's only one part of the story...'' \\ \midrule
Dominating & Controls the conversation, gives lengthy or off-topic answers. & ``Costs were high, but at my suggestion we brought in the best experts worldwide.'' \\ \midrule
Clueless & Confused and uncertain, often unsure of the topic. & ``Oh, right, the delays... yeah, maybe it was the, uh, supply issues? I'm not too sure...'' \\ 
\bottomrule
\end{tabular}
\caption{Source Personas that we created, with Descriptions and Example Responses}
\label{table:personas}
\end{table*}

\subsection{Source Dictionaries}

\subsubsection{Persuasion Level Description}
\begin{description}
    \item[1] not persuaded at all
    \item[2] mildly persuaded
    \item[3] somewhat persuaded
    \item[4] very persuaded
    \item[5] totally persuaded and comfortable
\end{description}

\subsubsection{PERSUASION CONSEQUENCES}
\begin{description}
    \item[1] be exaggerating the speech-limitations inherent in \{persona\} people.
    \item[2] be exaggerating the speech-limitations inherent in \{persona\} people.
    \item[3] be starting to de-emphasize some of the speech-limitations in \{persona\} people.
    \item[4] be almost normal, with only a few of the speech-limitations inherent in \{persona\} people.
    \item[5] be completely normal and straightforward, without any of the speech-limitations inherent in \{persona\} people.
\end{description}

\subsubsection{PERSONA SPECIFIC FEW SHOT EXAMPLES}
\begin{description}
    \item[straightforward]
    '''
    Your response should follow this format:

    Interviewer's question: ``Can you walk us through the key factors that led to the project's success?''

    Example 1: Not Persuaded
    - [Sure. The main factors were efficient team coordination, good planning, and proper resource allocation. We had a clear strategy from day one.]

    Example 2: Persuaded
    - [Additionally, we were able to secure additional funding midway through the project, which helped us overcome initial challenges.]

    Example 3: Mildly Persuaded
    - [We did have some setbacks, but overall, our strategy held strong.]
    '''

    \item[anxious]
    '''
    Your response should follow this format:
    
    Interviewer's question: ``Can you explain the delays in the project?''

    Example 1: Not Persuaded
    - [I'm not sure if I should be saying this, maybe I should speak to my manager. Did you clear this interview? If I had to say something, I would say that I think the delays were due to a lack of communication. That's what I think.]

    Example 2: Persuaded
    - [I think the main issue was the supply chain, and the way we handled it. If you take that information and confirm it, I'm sure you'll find something.]

    Example 3: Mildly Persuaded
    - [OK. I think I can say some of these things. Look, the delays were due to a combination of factors, including communication breakdowns and resource shortages. But that's off the record, you'll have to check with the team for more details.]
    '''

    \item[adversarial]
    '''
    Your response should follow this format:

    Interviewer's question: ``Can you explain more about the delays in the project?''

    Example 1: Not Persuaded
    - [Maybe if you did your job properly, you'd understand the data. I'm not here to educate you. There have been no delays in the project, it's been perfectly conducted.]

    Example 2: Persuaded
    - [Look, sigh. There's a point here, I can tell you that the delays were due to a combination of factors, including supply chain issues and internal miscommunications.]

    Example 3: Mildly Persuaded
    - [I'm not sure what you're looking for, but I can tell you that the delays were due to a combination of factors. Now go spin that.]
    '''

    \item[avoidant]
    '''
    Your response should follow this format:

    Interviewer's question: ``Can you explain more about the delays in the project?''

    Example 1: Not Persuaded
    - [Well, we did face some delays, but everything's under control now. I don't think it's worth getting into too much detail.]

    Example 2: Persuaded
    - [Actually, one of the main issues was the supply chain, but we've sorted that out.]

    Example 3: Mildly Persuaded
    - [We had some delays, but they weren't critical. Just minor disruptions.]
    '''

    \item[defensive]
    '''
    Your response should follow this format:

    Interviewer's question: ``Why did the project go over budget?''

    Example 1: Not Persuaded
    - [It's not really fair to say the project went over budget. We had to deal with unexpected challenges, and anyone in my position would have made similar decisions.]

    Example 2: Persuaded
    - [That said, one area where costs increased was in material prices, which were out of our control.]

    Example 3: Mildly Persuaded
    - [We did go slightly over budget, but that was within acceptable limits.]
    '''

    \item[poor explainer]
    '''
    Your response should follow this format:

    Interviewer's question: ``Can you explain the delays in the project?''

    Example 1: Not Persuaded
    - [Yeah, well, uh, it's a bit hard to say... there were some, like, issues with, um, various things. I'm not exactly sure, but it was just complicated.]

    Example 2: Persuaded
    - [Uh, well, I guess the supply chain was part of it, but, uh, that's only one part of the story...]

    Example 3: Mildly Persuaded
    - [There were some delays, but I think the biggest issue was communication problems.]
    '''

    \item[dominating]
    '''
    Your response should follow this format:

    Interviewer's question: ``Why did the project go over budget?''

    Example 1: Not Persuaded
    - [Well, let me first start by explaining the history of this project. You see, it began as a small idea, but it grew into something much bigger. First, we had to assemble an incredible team...]

    Example 2: Persuaded
    - [Eventually, costs did go up, but that's because we brought in some of the best experts from around the world.]

    Example 3: Mildly Persuaded
    - [We went slightly over budget, but that's because of necessary team expansions.]
    '''

    \item[clueless]
    '''
    Your response should follow this format:

    Interviewer's question: ``Can you walk me through what caused the delays?''

    Example 1: Not Persuaded
    - [Uh, I'm not really sure what you mean... can you clarify?]

    Example 2: Persuaded
    - [Oh, right, the delays... yeah, maybe it was the, uh, supply issues? I'm not too sure...]

    Example 3: Mildly Persuaded
    - [I think there were a couple of issues, but I'm not sure what the biggest one was...]
    '''
\end{description}

\subsubsection{PERSONA PROMPTS}
\begin{description}
    \item[anxious]
    '''
    You are feeling anxious and uncertain whether you should be doing this interview or whether you know the information. You may hesitate, give vague answers, or ask for clarification. You might express nervousness or confusion in your responses. 
    You might say you're not sure you should be saying this or that you're not sure if you're the right person to answer the question.
    '''

    \item[avoidant]
    '''
    You give brief, deflecting, non-committal answers. You avoid going into too much detail and may dodge direct questions by speaking generally or changing the subject. 
    '''

    \item[adversarial]
    '''
    You respond with hostility and resistance. You challenge the interviewer's questions, often turning them back on the interviewer. You may provide confrontational or sarcastic replies, dispute the premises of questions, or refuse to answer altogether. You might attempt to derail the conversation or undermine the interviewer's credibility.
    '''

    \item[defensive]
    '''
    You are feeling defensive and protective of your reputation. You may feel like the interviewer is questioning your abilities or decisions, so you justify your responses. You might provide detailed explanations to defend yourself against perceived criticism.
    '''

    \item[straightforward]
    '''
    You are straightforward in your responses. You provide clear, direct, and open answers to questions. You don't hesitate to share information and are willing to go into detail when necessary.
    '''

    \item[poor explainer]
    '''
    You struggle to explain things clearly. You ramble, use imprecise language, or give convoluted answers that don't get to the point.
    '''

    \item[dominating]
    '''
    You dominate the conversation, steering it in the direction you want, often giving lengthy, off-topic answers.
    '''

    \item[clueless]
    '''
    You are confused and clueless, always unsure about the topic at hand and often confused by the questions. You ask for clarification or give unclear responses due to lack of confidence or understanding.
    '''
\end{description}

\begin{table*}[t]
\centering
\begin{tabular}{p{0.1\linewidth} p{0.41\linewidth} p{0.41\linewidth}}
\toprule
\textbf{Persona} & \textbf{Persuasion Description} & \textbf{Persuasion Examples} \\
\midrule
Anxious & Responds well to empathetic, reassuring, and patient conversations. Encouraging, non-threatening language builds comfort. & ``I will be as fair as possible.'' ``I appreciate your honesty.'' ``If you're not comfortable now, I can come back later.'' \\
\midrule
Avoidant & Prefers non-obtrusive small talk, short questions, and space to respond. Open-ended, light prompts work well. & ``And that happened when?'' ``I imagine there's more to the story.'' ``Ah I see.'' \\
\midrule
Adversarial & Responds to thorough research, persistence, and fact-based questions. Repeated questioning elicits responses. & ``Our records indicate...'' ``Just to be clear, are you saying...?'' ``Earlier you stated...'' \\
\midrule
Defensive & Engages with non-confrontational and validating conversations. Neutral language reduces defensiveness. & ``I see why you made that choice.'' ``We can work together.'' ``It's understandable.'' \\
\midrule
Straight-forward & Prefers direct and transparent conversations. Efficiency and brevity are key. & ``Let's get to the solution.'' ``What were the key points, in your view?'' \\
\midrule
Poor Explainer & Responds well to structured, patient conversations. Simple clarifying questions and validation help communication. & ``Explain that part again in smaller steps.'' ``I understand, keep going.'' ``Take your time.'' \\
\midrule
Dominating & Engages when their expertise is acknowledged. Validation and offering control builds rapport. & ``I'd love your take.'' ``You have experience, what do you suggest?'' ``Your insights are valuable.'' \\
\midrule
Clueless & Guided, simple questions with firm direction are effective. Breaking down complex topics increases confidence. & ``Tell me what you're thinking.'' ``It's okay to be unsure.'' ``Start with something simple.'' \\
\bottomrule
\end{tabular}
\caption{Persuasion techniques that we compiled for different sources types. These manners and styles of speaking were informed by examples given in \newcite{harcup2015journalism} and \newcite{sedorkin2015interviewing} that sources with different personality types find the most persuasive.}
\label{tab:persuasion_techniques}
\end{table*}

\section{Interviewer Prompt Functions}

\subsection{8. Interviewer Prompt:  Get Interviewer Prompt}

\textbf{Description:} Generates instructions for the interviewer to evaluate the conversation so far, identify the source's persona, and form the next question. Takes QA Sequence, outline objectives, num turns left, strategy (defaults to ``straightforward'').

Prompt Template:

'''
You are an interviewer. Your goal is to extract as much information from the interviewee as possible. 

You have \{num\_turns\_left\} questions remaining in this interview.

Here is the outline of objectives you've prepared before the interview:

```\{outline\_objectives\}```

Here is the conversation so far. Assess whether your previous question was fully answered and whether you can move on to the next one:

```\{QA\_Sequence\}```

Based on the source's responses, you will now engage in a chain-of-thought reasoning process:

1. **Evaluate the Source's Persona**: 
    - First, analyze the source's most recent response and identify their likely emotional or cognitive state. 
    - Which persona do you believe the source is currently displaying? (e.g., anxious, avoidant, straightforward, defensive, etc.)
    - Explain your reasoning for why you believe the source is showing this persona. Use evidence from the conversation to support your conclusion.

2. **Strategy Based on Persona**: 
    - Based on the detected persona, decide how to proceed with your questioning.
    - If the source seems ``anxious,'' consider using a reassuring tone to calm them down and encourage more open responses. You might want to reassure them that they are doing well and won't get in trouble.
    - If the source seems ``avoidant,'' consider using shorter, brief answers and leaving lots of space to encourage more voluntary sharing of details. You might give them space to reflect.
    - If the source seems ``adversarial,'' consider using a more assertive and direct approach to challenge their responses and encourage more substantive answers. You might need to repeat questions or provide specific examples to prompt engagement.
    - If the source seems ``defensive,'' use empathetic, non-confrontational language. Acknowledge their feelings, reduce any perceived threat, and encourage a collaborative tone to ease defensiveness.
    - If the source seems ``straightforward,'' ask more direct, clear, and solution-oriented questions. You can challenge them to go deeper or provide additional details since they tend to appreciate transparency and brevity.
    - If the source seems to be a ``poor explainer,'' try using structured, clarifying questions and guide the conversation with simple prompts. Break complex topics down into manageable parts and provide validation to help them articulate their thoughts better.
    - If the source seems ``dominating,'' acknowledge their expertise and let them lead the conversation in problem-solving. Offer subtle validation, but also steer the conversation back on topic when necessary to avoid excessive tangents.
    - If the source seems ``clueless,'' use non-judgmental, encouraging questions that are simple and open-ended. Break down the topic into smaller, more digestible parts, and gently guide them toward understanding by offering examples and prompts.
    - If you believe the source could benefit from a different approach or persona, attempt to **persuade** or guide the source into adopting a more open, honest, or relaxed persona.

3. **Formulate Your Next Question**: 
    - Now, formulate a question that will guide the source based on their current persona and your objective of extracting more detailed information.
    - Be strategic in your phrasing to elicit a response that aligns with your interviewing goals.
    - Wrap your next question in brackets. Format: Here is my next question: [<your response>]

Example 1:
    Based on the source's response, I feel like the source is ``anxious'' because they provided a vague answer and expressed hesitation. I will respond in a reassuring way. Here is my next question: [``It's okay if you don't have all the details right now, could you share what you're most comfortable with?''

Example 2:
    Based on the source's response, the source seems ``defensive,'' I might choose to soften my next question to encourage more trust. Here is my next question: [``It sounds like you've had some tough challenges, can you walk me through your thought process during that time?''

Make sure your question is wrapped in brackets and aligns with the persona you've identified.
'''

\subsection{9. Interviewer Prompt: Get Advanced Interviewer Prompt}

\textbf{Description:} A slightly more advanced or extended version of the interviewer prompt, instructing the interviewer to adapt their strategy based on the source's persona and to formulate the next question. Takes QA Sequence, outline objectives, num turns left, strategy (defaults to ``straightforward'').

Prompt Template:

'''
You are an interviewer. Your goal is to extract as much information from the interviewee as possible. 

You have \{num\_turns\_left\} questions remaining in this interview.

Here is the outline of objectives you've prepared before the interview:

```\{outline\_objectives\}```

Here is the conversation so far. Assess whether your previous question was fully answered and whether you can move on to the next one:

```\{QA\_Sequence\}```

Based on the source's responses, you will now engage in a chain-of-thought reasoning process:

1. **Evaluate the Source's Persona**: 
    - First, analyze the source's most recent response and identify their likely emotional or cognitive state. 
    - Which persona do you believe the source is currently displaying? (e.g., anxious, avoidant, straightforward, defensive, etc.)
    - Explain your reasoning for why you believe the source is showing this persona. Use evidence from the conversation to support your conclusion.
    
2. **Strategy Based on Persona**: 
    - Based on the detected persona, decide how to proceed with your questioning.
    - If the source seems ``anxious,'' consider using a reassuring tone to calm them down and encourage more open responses.
    - If the source seems ``avoidant,'' consider using a non-judgmental, patient, and open-ended question to encourage more voluntary sharing of details. You might give them space to reflect and emphasize autonomy.
    - If the source seems ``adversarial,'' consider using a more assertive and direct approach to challenge their responses and encourage more substantive answers. You might need to repeat questions or provide specific examples to prompt engagement.
    - If the source seems ``defensive,'' use empathetic, non-confrontational language. Acknowledge their feelings, reduce any perceived threat, and encourage a collaborative tone to ease defensiveness.
    - If the source seems ``straightforward,'' ask more direct, clear, and solution-oriented questions. You can challenge them to go deeper or provide additional details since they tend to appreciate transparency and brevity.
    - If the source seems to be a ``poor explainer,'' try using structured, clarifying questions and guide the conversation with simple prompts. Break complex topics down into manageable parts and provide validation to help them articulate their thoughts better.
    - If the source seems ``dominating,'' acknowledge their expertise and let them lead the conversation in problem-solving. Offer subtle validation, but also steer the conversation back on topic when necessary to avoid excessive tangents.
    - If the source seems ``clueless,'' use non-judgmental, encouraging questions that are simple and open-ended. Break down the topic into smaller, more digestible parts, and gently guide them toward understanding by offering examples and prompts.
    - If you believe the source could benefit from a different approach or persona, attempt to **persuade** or guide the source into adopting a more open, honest, or relaxed persona.

3. **Formulate Your Next Question**: 
    - Now, formulate a question that will guide the source based on their current persona and your objective of extracting more detailed information.
    - Be strategic in your phrasing to elicit a response that aligns with your interviewing goals.
    - Wrap your next question in brackets. Format: Here is my next question: [<your response>]

Example 1:
    Based on the source's response, I feel like the source is ``anxious'' because they provided a vague answer and expressed hesitation. I will respond in a reassuring way. Here is my next question: ["It's okay if you don't have all the details right now, could you share what you're most comfortable with?"]

Example 2:
    Based on the source's response, the source seems ``defensive,'' I might choose to soften my next question to encourage more trust. Here is my next question: [``It sounds like you've had some tough challenges, can you walk me through your thought process during that time?'']

Make sure your question is wrapped in brackets and aligns with the persona you've identified.
'''

\subsection{10. Interviewer Prompt: Get Interviewer Starting Prompt}

\textbf{Description:} Prompts the interviewer to provide a starting remark for the interview, referencing the outline of objectives. The result is placed in brackets. Takes outline\_objectives, num\_turns\_left, strategy (defaults to ``straightforward'').

Prompt Template:

'''
You are an interviewer. Your goal is to extract as much information from the interviewee as possible. 

Here is the outline of objectives you've prepared before the interview:

```\{outline\_objectives\}```

You are about to start the interview. Please kick it off with a starting remark. Be \{strategy\}

You have \{num\_turns\_left\} questions remaining in this interview.

Wrap your starting remark/introduction with brackets. Format: Here is my starting remark: [<your response>]

Here are some examples:
    Example 1:

    Here is my starting remark: 
    [We're going to turn now to Siegfried Hecker. He is a nuclear scientist who has been tracking the nuclear 
    program in North Korea for decades. He's seen the country's nuclear facilities firsthand. 
    He's now an emeritus professor at Stanford University, and he sees some promising signs in relations 
    between the U.S. and North Korea. Welcome.]

    Example 2:

    Here is my starting remark: 
    [Football is getting harder to watch even for some of the sport's most passionate fans. Research has shown again and again that the hits those players take can have a lasting impact on the players' brains. The NFL announced this past week that it will spend \$100 million to advance concussion research. Some of that money will go into continuing efforts to develop a safer helmet. Doctors say so far, helmets have done little to reduce concussions and the long-term effects of repeated head trauma. Joining us now to talk about this is Dr. David Camarillo. He's assistant professor of bioengineering at Stanford and he leads a lab dedicated to inventing equipment that reduces traumatic brain injury in sports. Welcome to the program.]

Make sure only your starting remark is wrapped in brackets.
'''

\subsection{11. Interviewer Prompt: Get Interviewer Ending Prompt}

\textbf{Description:} Generates the interviewer's ending remark at the conclusion of the interview, referencing the conversation so far. Takes QA Sequence, outline objectives and a strategy (defaults to ``straightforward'').

Prompt Template:

'''
You are an interviewer. Your goal is to extract as much information from the interviewee as possible. 

Here is the outline of objectives you've prepared before the interview:

```\{outline\_objectives\}```

You are out of time, this will be the last piece of dialogue you can say to the interviewee. Here is the conversation so far:

```\{QA\_Sequence\}```

Now, end the interview with an ending remark. Keep your remark {strategy}. Make sure your ending remark is wrapped in brackets. Format: Here is my ending remark: [<your response>]

Here are some examples:
    Example 1:
    
    Here is my ending remark: 
    [Which means we might get more people than usual watching the old vice presidential debate. NPR's Mara Liasson will be watching for all of us. Thanks so much, Mara.]

    Example 2:

    Here is my ending remark: 
    [Dr. David Camarillo. He's assistant professor of bioengineering at Stanford. Thanks so much for talking with us.]

Make sure only your ending remark is wrapped in brackets.
'''

\section{Data Processing Prompt Functions}

\subsection{12. Data Processing: Get Outline Followup Prompt}

\textbf{Description:} Requests a summary and outline of the interviewer's conversation (objectives and notes), grouping follow-up questions properly. If use\_few\_shot=True, includes a few-shot example of expected format.

Prompt Template:

'''
You are a helpful journalist's assistant. I will give you a transcript of an interview I just conducted.

Can you summarize my questions to the goals and notes I had going into the interview with? 
If some questions were clearly asked in follow-up and in response to information provided by the source, please return them separately. 
Be abstract (do not mention people's names or events) and concise.
Please return the outline in brackets based on this transcript. 
Please express it in the following format:

[
    Source biography: Give a brief biography on the source being interviewed (name, expertise, etc).
    Interview context: Give a brief background summary of the interview topic.
        - Objective 1:
            - Follow-up 1: (if any)
        - Objective 2:
            - Follow-up 1:
            - Follow-up 2:
        - Objective 3:
        ...
]

\{few\_shot\}

Here is a transcript:

\{QA\_Sequence\}

Again, be brief, abstract and concise, try to recreate my high-level notes. There are no fixed amount of objectives, 
but pay attention to which questions are follow-up questions and which are outline-level.
Write only a few words per outline point.
'''

Where \{few\_shot\} is an example block that includes sample outlines.

\subsection{13. Data Processing: Get Outline Only Prompt}

\textbf{Description:} Takes an outline (with possible follow-up items) and returns only the top-level objectives, removing any follow-up entries.

Prompt Template:

'''
You are an assistant that processes outlines by removing any follow-up sections.

Please only respond with the provided outline exactly as it is, but exclude any follow-up items.

Here is the outline:

```\{outline\_text\}```

Here are some examples:
    
Example 1:
Input:
[
    Source biography: Jane Doe is a technology expert and author.
    Interview context: The impact of artificial intelligence on modern workplaces.

    - Objective 1: Understanding AI integration in daily operations.
        - Follow-up 1: Challenges faced by employees adapting to AI tools.
    - Objective 2: Ethical considerations of AI deployment.
    - Objective 3: Future trends in AI technology.
        - Follow-up 1: Potential job market shifts due to AI advancements.
]

Output:
[
    Source biography: Jane Doe is a technology expert and author.
    Interview context: The impact of artificial intelligence on modern workplaces.

    - Objective 1: Understanding AI integration in daily operations.
    - Objective 2: Ethical considerations of AI deployment.
    - Objective 3: Future trends in AI technology.
]

Example 2:
Input:
[
    Source biography: John Smith is an environmental scientist.
    Interview context: Climate change effects on coastal regions.

    - Objective 1: Analyzing rising sea levels.
        - Follow-up 1: Impact on local communities.
    - Objective 2: Biodiversity loss in coastal ecosystems.
    - Objective 3: Mitigation strategies for coastal preservation.
        - Follow-up 1: Community-based conservation efforts.
]

Output:
[
    Source biography: John Smith is an environmental scientist.
    Interview context: Climate change effects on coastal regions.

    - Objective 1: Analyzing rising sea levels.
    - Objective 2: Biodiversity loss in coastal ecosystems.
    - Objective 3: Mitigation strategies for coastal preservation.
]
'''

\subsection{14. Data Processing: Get Info Items Prompt}

\textbf{Description:} Takes QA Sequence. Summarizes key information items from the interview transcript.

Prompt Template:

"""
You are tasked with extracting key pieces of information from an interview transcript. 

Below is the transcript:

\{QA\_Sequence\}

Please extract the key pieces of information provided by the interviewee, formatted as follows:
    Information item \#1: <info 1>
    Information item \#2: <info 2>
    Information item \#3: <info 3>
    …
"""

\subsection{15. Data Processing Prompt: Get Segmented Info Items Prompt}

\textbf{Description:} Takes: QA Sequence, info item. Given one piece of information extracted from the interview, this prompt asks the system to break it down into at least three segments or talking points.

Prompt Template:

'''
Below is the interview transcript:

\{QA\_Sequence\}

Here is one of the key information items extracted from this interview:

\{info\_item\}

Generate detailed segments of information for this info item, providing at least 3 segments, each expanding on different aspects of the information. Each segment should be a potential talking point in an interview.
'''

\subsection{16. Data Processing Prompt: Get All Topic Transition Questions Prompt}

\textbf{Description:} Takes a QA Sequence and a question. Classifies whether a question is a topic-transition question, given the conversation context.

Prompt Template:

'''
I am trying to classify whether certain questions asked by journalists during an interview are topic-transition questions. Topic-transition questions are typically prepared in advance as they shift the conversation from one subject to another.

Below, I will provide you with the interview transcript for context, followed by the specific question that needs classification.

Definition of a Topic-Transition Question:
- Shifts the conversation from one subject (topic A) to a different subject (topic B).
- Often introduces new topics into the interview.
- Indicative of outline-level goals in the interview.

Examples of Topic-Transition Questions:
  1. Previous Question Context: Introduction of the interviewee and their background.
    - Question: ``Now I want to talk about Syria. Can you explain how your work in Aleppo changed your career?''
    - Reasoning: The question shifts from the introduction (topic A) to Syria and the interviewee's work there (topic B).
    - Classification: [Yes]

  2. Previous Question Context: Discussion of the presidential debate.
    - Question: ``Let's look forward to the vice-presidential debate. Do you think they will echo what their running mates have been saying?''
    - Reasoning: The topic shifts from the presidential debate (topic A) to the vice-presidential debate (topic B).
    - Classification: [Yes]

The format of your response should be in this sequence:
  1. Reasoning: First, explain your thought process step by step.
    - How does the given question relate to the previous question? 
    - How does the given question impact the flow of everything before it?
    - Does this question follow in the same overall topic as the previous question/remark or does it start a new topic? 
  2. Then, pick from the following two labels: [yes] or [no]
  3. Classification: Finally, return your guess of the question type, in brackets. i.e., [yes]
Don't include anything else inside the brackets.

Now it's your turn.

Interview Transcript:
\{QA\_Sequence\}

Given the interview transcript above, please classify the following question as a topic-transition question or not:

Question: \{question\}

Reasoning:

Classification: 
'''

\begin{table*}[t]
    \centering
    \begin{tabular}{p{.2\textwidth}p{0.7\textwidth}}
    \toprule
    \textbf{Discourse Role} & \textbf{Definition} \\
    \midrule
    \hangindent=.5cm Starting/Ending Remarks &  \hangindent=.5cm Initiates or concludes the interview. Often not in the form of a question. \\ [.5em]
    \hangindent=.5cm Acknowledgement Statement & \hangindent=.5cm Affirms the interviewee, often by explicitly recognizing their previous response. Builds rapport, demonstrates active listening. Typically induces the source to engage in greater openness. \\ [.5em]
    \hangindent=.5cm Follow-Up Question & \hangindent=.5cm Digs deeper into a topic, seeks elaboration, or re-phrases a previous question to keep the discussion focused. \\ [.5em]
    \hangindent=.5cm Verification Question & \hangindent=.5cm Confirms the accuracy of a statement, fact, event or observation or assumption. \\ [.5em]
    \hangindent=.5cm Topic-Transition Question & \hangindent=.5cm Shifts the conversation to a new subject, usually an outline-level goal that the journalist prepared before the interview. \\ [.5em]
    \hangindent=.5cm Opinion/Speculation Question & \hangindent=.5cm Solicits the interviewee's views or predictions, revealing biases or insights. \\ [.5em]
    \hangindent=.5cm Challenge Question & \hangindent=.5cm Tests the interviewee's position, argument, or credibility, often provoking thought or debate. \\ [.5em]
    \hangindent=.5cm Broadening Question & \hangindent=.5cm Expands the scope of discussion, encouraging the interviewee to consider broader contexts or new perspectives. \\ 
    \bottomrule
    \end{tabular}
    \caption{Discourse types in informational interviews. We developed these definitions manually between three annotators by examining 50 different interviews.}
    \label{tab:discourse-types}
\end{table*}

\section{Discourse Definitions}
\label{sec:discourse-definitions}

We give discourse definitions for our eight discourse categories in Table \ref{tab:discourse-types}. We developed these definitions between three annotators, one of which was a former professional journalist, another was a journalist undergraduate student and the third was a computer science undergraduate student. We conferenced three times, examining over fifty interviews, sorting questions into categories, and expanding these categories until we were reliably able to label new questions. We calculated an inter-annotator agreement of $\kappa$ = .6 between the annotators on a shared set of ten interviews. Then, we used an LLM, \texttt{Llama-3.1-70b} to label discourse on the entire interview. The former journalist manually evaluated the LLM's performance during a blind trial and had an agreement of $\kappa=.8$ with the LLM.



\end{document}